\documentclass[a4paper,conference]{IEEEtran}

\usepackage{array}
\usepackage{url}
\usepackage{xcolor}
\usepackage{subcaption}
\usepackage{graphicx}
\usepackage{amsmath}
\usepackage{amssymb}
\usepackage{multirow}
\usepackage{booktabs}
\usepackage{cite}
\usepackage{url}

\begin{document}
%
\title{Map-Based Temporally Consistent Geolocalization \\ through Learning Motion Trajectories}


\author{
\IEEEauthorblockN{Bing Zha \quad Alper Yilmaz}
\IEEEauthorblockA{Photogrammetric Computer Vision Laboratory \\
The Ohio State University\\
Columbus, OH, USA \\
Email: \{zha.44, yilmaz.15\}@osu.edu}
}

\maketitle

\begin{abstract}
In this paper, we propose a novel trajectory learning method that exploits motion trajectories on topological map using recurrent neural network for temporally consistent geolocalization of object. Inspired by human's ability to both be aware of distance and direction of self-motion in navigation, our trajectory learning method learns a pattern representation of trajectories encoded as a sequence of distances and turning angles to assist self-localization. We pose the learning process as a conditional sequence prediction problem in which each output locates the object on a traversable path in a  map. Considering the prediction sequence ought to be topologically connected in the graph-structured map, we adopt two different hypotheses generation and elimination strategies to eliminate disconnected sequence prediction. We demonstrate our approach on the KITTI stereo visual odometry dataset which is a city-scale environment and can generate trajectory with metric information. The key benefits of our approach to geolocalization are that 1) we take advantage of powerful sequence modeling ability of recurrent neural network and its robustness to noisy
input, 2) only require a map in the form of a graph and simply use an affordable sensor that generates motion trajectory and 3) do not need initial position. The experiments show that the motion trajectories can be learned by training an recurrent neural network, and temporally consistent geolocation can be predicted  with both of the proposed strategies. 
\end{abstract}


%
\IEEEpeerreviewmaketitle

\section{Introduction}

One of the enduring challenges in robotics and self-driving cars is to localize themselves in the environment. Yet
human brain as a brilliant information processor is exceptionally good at self-localization and navigation in a known environment even without any visual cues. Such extraordinary ability have attracted much attention from neuroscientists and philosophers to look for explanations and models on how the brain performs these two fundamental tasks. A few early neuroscience studies have shown that an internal map of the environment and a 
navigation strategy are both needed to locate ourselves and navigate toward a goal \cite{OKEEFE1971171}, \cite{OKEEFE197678}, \cite{erdem2012goal}. Understanding such process and building computational models are crucial and necessary to introduce such functionality in advanced artificial intelligence applications, such as robotics and self-driving cars. 

In parallel to the exploration of biological mechanism for self-localization and navigation, researchers have also developed other techniques to achieve these functionalities. The most popular of these engineered  techniques is the Global Positioning System (GPS) that was established around 1970s and uses an external network of satellites that continuously transmit time-coded information \cite{hofmann2012global}. This information is received and interpreted by a receiver to precisely estimate the location of the object. 
Although this technology has been popular in many location-based applications, the limitation is still notable: for instance GPS signal might be unavailable or unreliable under the ground, in tunnels and indoors. 

Unlike the GPS embedded in devices, our brain's system accesses location and navigation information by integrating multiple signals relating internal self-motion and external landmarks. Recent research \cite{hafting2005microstructure}, \cite{bush2015using}, \cite{edvardsen2020navigating} has shown that the mammalian brain uses an incredibly sophisticated GPS-like localization and tracking system of its own to help recognize location and guide us from one location to the next. This ability to figure out where we are and where we need to go toward a goal, as one of the most fundamental survival skills, is extremely significant for animal and human. One typical method used in animals, such as desert ants and honeybees, is called \textit{path integration}, a mechanism in which neurons calculates location simply by integrating self-motion information including direction and speed of movement - a task carried out without reference to external cues such as physical landmarks \cite{bush2015using}. Another method suggested to represent space as a graph structure in which nodes denote specific places and links are formed between pairs of place. The resulting graph reflects the topology of the explored environment upon which localization and navigation can be directly implemented by graph search algorithm \cite{edvardsen2020navigating}. 


Neuroscientists' explanation of human brain suggests that the way we perform localization and navigation is different from the approaches taken by engineers \cite{tolman1948cognitive}, \cite{OKEEFE1971171}, \cite{OKEEFE197678}, \cite{hafting2005microstructure}. Their research shows localization and navigation in mammals depend on a distributed, modularly organized brain network which computes and represents positional and directional information, and it also shows the brain contains a directionally oriented, topographically organized neural map of the spatial environment \cite{hafting2005microstructure}. 

Inspired by those pioneering research of animal's localization and navigation conducted in neuroscience field, we introduce a topological map-based trajectory learning method that infers sequential geolocations of object through training different motion trajectories sampled from topological map. Compared to \textit{path integration} method which takes velocity as input and metric coordinates as output that prone to be sensitive to noise and drift, our approach feeds in network discrete turning angle as input and graph edge id as output. Moreover, the topological place graph structure is used to locate object coarsely which tend to be the way of geolocalization human perform. Lastly, in order to address temporally and topologically consistent geolocalization, we introduce two different hypotheses generation and pruning strategies to ensure connectivity of output edge location. We test our approach on the KITTI \cite{geiger2012we} stereo visual odometry dataset which consists of stereo images of city rural area. We show that our method can sequentially locate object based on trained network. 

The key contributions of this paper are as follows:
\begin{itemize}
    \item We introduce a novel topological map-based trajectory learning method for geolocalization of object which can be consider as a mixture of \textit{path integration} and place graph method.
    \item We introduce two hypotheses generation and pruning strategies for temporally and topologically consistent geolocalization on graph.
\end{itemize}

The remainder of the paper is organized as follows. In  Section \ref{sec:related_work},  related works of geolocalization from engineering and neuroscience field are summarized. Section \ref{sec:method} presents building blocks of the proposed approach, which is divided into three parts: trajectory learning and geolocalization, temporal consistency, and trajectory generation. Experiments and results of trajectory learning and consistent geolocalization are shown in Section \ref{sec:experi}. We conclude the paper in Section \ref{sec:con}.  

\section{Related Work}
\label{sec:related_work}
In order to access our contribution in relation to the vast literature on localization and navigation problem, it is important to consider what constraints and assumptions these approaches have used and what limitations exist or whether these approaches make use of the benefits of neural network as a tool. Below, we summarize several types of methods that are representative and related to ours.  


\textbf{SLAM and Neural SLAM.} \enspace Classical method such as Simultaneous Localization and Mapping (SLAM) \cite{cadena2016past} can construct the map of an unknown environment while simultaneously keeping track of the object's location on the map that is being generated. Although there are many efficient SLAM variants that use different sensors and different map representations such as metric \cite{30720}, topology \cite{boal2014topological}  or hybrid \cite{tomatis2003hybrid}, they heavily depend on the sensor setup, undergo intensive computation to fuse the information and suffer from drift errors. Another neural SLAM system was presented in \cite{milford2004ratslam}, \cite{milford2010persistent} to mimic rodent's navigation behavior in which path integration and attractor network were used. However, our approach avoids path integration step due to drift problems and only uses distance and direction information which can be robust to noisy sensor measurements.  

\textbf{Visual Localization.} \enspace A major category of work in the literature is dedicated to the use of images for localization referred to as \textit{visual localization}. These methods can be  classified into photogrammetric localization \cite{sattler2011fast} and image matching based localization \cite{walch2017image}. Both of these methods are computationally expensive and require large databases of geotagged images which is hard to build in large scales and maintain to mitigate changes in the environment. In general, all the techniques  given above are either highly dependent on appearance of the environment or external information for localization which considerably restrict the usability and accessibility.

\textbf{Probabilistic Localization.} \enspace A common form of localization problem is to use sensory readings to estimate absolute coordinates of the object and its uncertainty on the map using probability theory. The authors of \cite{fox1999monte} presented a Markovian approach to model the posterior distribution of an object's position given its pose. The recent work \cite{brubaker2015map} proposed a probabilistic self-localization method using OpenStreetMap and visual odometry. The authors of \cite{gupta2016gps} presented an localization approach based on stochastic trajectory matching using a brute-force search. However, all of those methods require generating and maintaining posterior distributions which lead to complicated inference strategies and high computational cost. 

\textbf{Map Matching.} \enspace In commercial solutions, the task of geolocalization is often approached by a \textit{map matching} technique which relates an external position signal, such as GPS, to a road network \cite{newson2009hidden}. The methods in this category are typically based on hidden markov model and are able to localize the object by matching the GPS trajectory to the road graph. However, they explicitly require an external georeferenced signal and cannot be scalable to cases when these signals are degraded or not available.

\textbf{Navigation in Neuroscience.} \enspace In neuroscience, several studies study the mechanism of animal's ability in mapping, localization and navigation \cite{tolman1948cognitive}, \cite{OKEEFE1971171}, \cite{OKEEFE197678}. In those ground-breaking studies, researchers have shown how the biological systems enable a sense of place and navigation based on specialized neurons (place cell \cite{OKEEFE197678}, grid cell \cite{hafting2005microstructure} and border cell \cite{barry2006boundary}, etc.) for each task. 
These discoveries show that navigation in mammals is thought to be supported by a ``cognitive map'' \cite{tolman1948cognitive}, an internal neural representation of environment. Such map would assist animal with navigational planning and self-localization. In \cite{edvardsen2020navigating}, author suggested that place cells are associated with a topological strategy for navigation, while grid cells are suggested to support metric vector navigation. Our approach is largely inspired by those studies and can be considered as a mixture of these two mechanisms. 

\textbf{Localization from Deep Learning.} \enspace There are a few similar studies closely related to us using deep learning \cite{wei2019localization}, \cite{zha2019localization} .
In \cite{wei2019localization}, a sequence to sequence labeling method for trajectory matching using neural machine translation network is proposed. This approach was shown to only work well on synthetic scenarios where the input trajectory was synthetically generated with known sequence of nodes from the map. \cite{zha2019localization} presented a variable length sequence classification method for motion trajectory localization in which temporal consistency of estimated locations were not guaranteed. In contrast, this paper introduces a topological map-based trajectory learning method and utilizes two different hypotheses generation and pruning strategies to consistently achieve geolocalization. 

\section{Proposed Method}    
\label{sec:method}  
Our method is built on the motivation that humans are exceptionally good at finding \textit{where they are} based on observations and a simple mind map. Our goal is to infer the location on a map through distance one has traversed and corners one has turned which leads to utilize the topological map represented in some form in the brain.
The first mental map concept was published in 1940's \cite{tolman1948cognitive} in which author proposed the ``cognitive map'' to indicate an internal neural space representation. 
Adopting this inspiring body of work, we develop a topological map-based approach using motion trajectory as training data to achieve self-localization. Formally, given a sequence of distances and turning angles along with a map, we sequentially predict which edge is located. In fact, the movement of object on graph is physically constrained by the condition of the topology of an underlying road network. In order to guarantee the connectivity of edge prediction, we adopt hypotheses generation and pruning strategy. Moreover, due to distance information needed, our approach requires absolute scale of motion trajectory which directly leads us to employ stereo visual odometry method to obtain motion trajectory.  

The following sections describe how this process is done. Firstly, we discuss our proposed topological map-based geolocalization process through trajectory learning including map representation, motion trajectory representation and the neural network architecture. Then, we introduce two different strategies to deal with temporal consistency of location. Lastly, we briefly present a visual odometry method on how to obtain metric motion trajectory. 

\subsection{Trajectory Learning and Edge Localization}

We develop a mathematical model that encodes biological system's treatment of spatial information in the topological map for navigation and geolocalization. The model uses recurrent neural network (RNN) architecture that learns object's motion trajectory while encoding the map. Our trajectory learning process utilizes the map in the form of a sequence of discrete distances and directions of object's motion. Formally, the RNN is trained to solve for the output edge location probability using the map and motion sequence: 
\begin{equation}
    P_{loc} = P(s_t | {\varphi, \beta}_{1:t}, M)
\end{equation}
where, $s_t$ is the output edge id, and $P_{loc}$ indicates the output result which is conditioned on the topological map,  $M$, and the sequence of direction and distance ${\varphi, \beta}_{1:t}$ where $t$ is the length of sequence.

\begin{figure}[t]
\centering
\begin{subfigure}[h]{0.24\textwidth}
   \includegraphics[width=4.2cm, height=4.2cm]{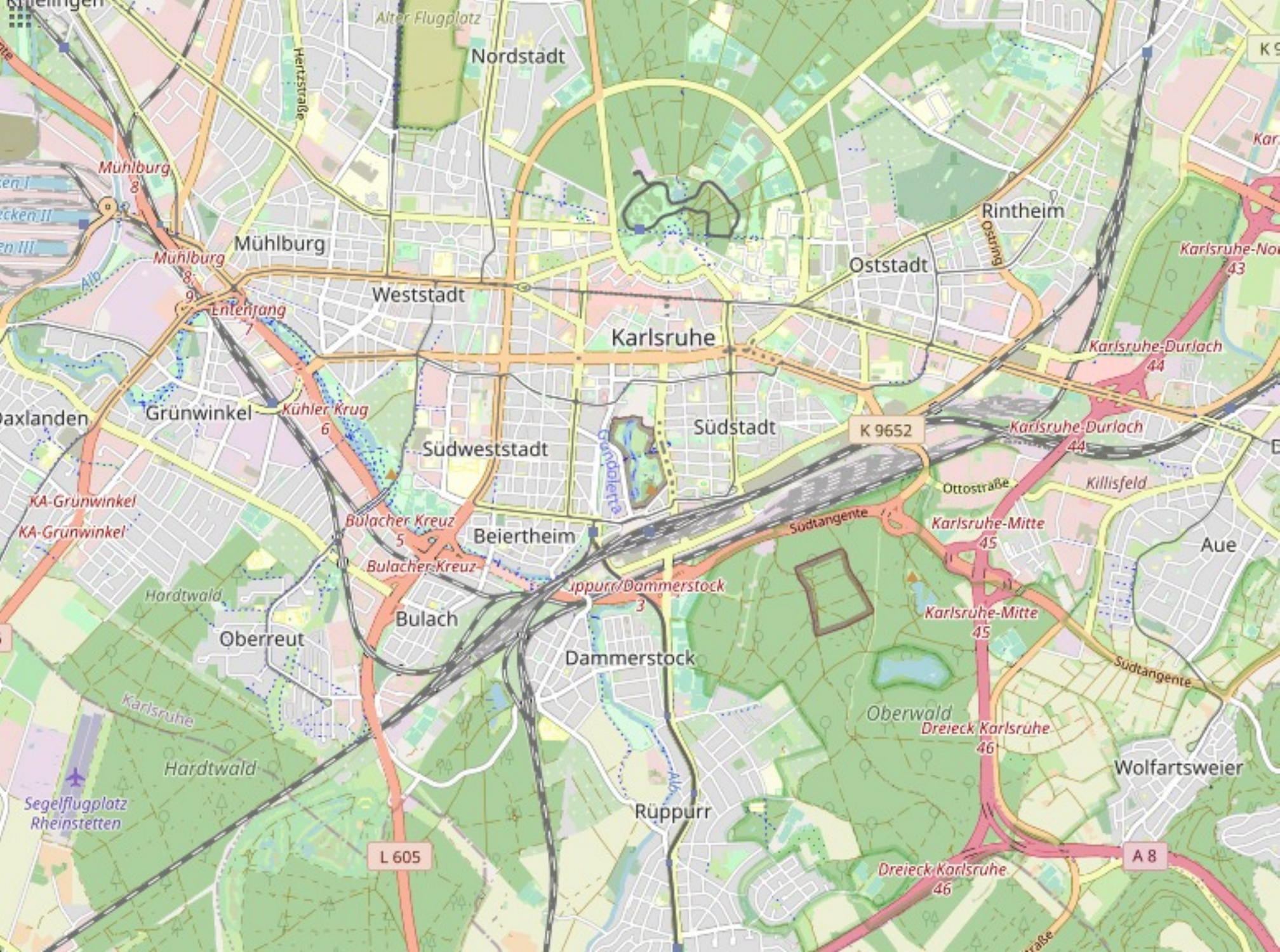}
   \caption{OSM map}  
   \label{fig:osm}  
\end{subfigure}  
\begin{subfigure}[h]{0.24\textwidth}  
   \includegraphics[width=4.2cm, height=4.2cm]{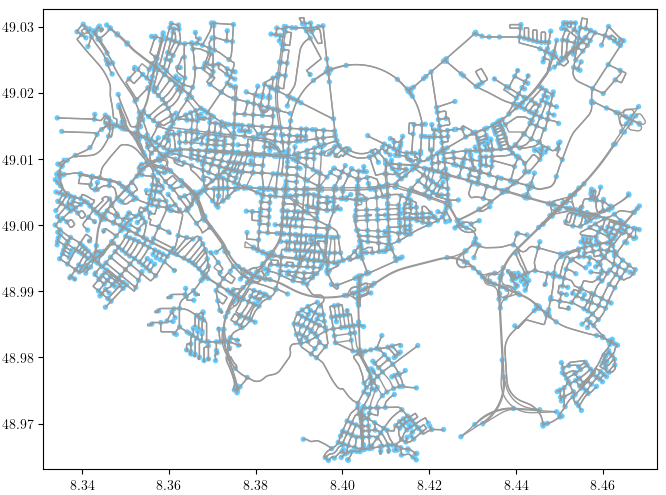}  
   \caption{Graph-structured map}  
   \label{fig:graph}  
\end{subfigure}     
\caption{OpenStreetMap representation. Left image shows the OSM map of area in Karlsruhe. The right image shows the graph generated  from the OSM map. In the graph the blue nodes and green edges represent places and roads respectively.}
\label{fig:osm_and_graph}
\end{figure}

Since our approach depends on the topological structure of a map for geolocalization, we use the crowd-sourced OpenStreetMap (OSM)\footnote{\url{https://www.openstreetmap.org}} from which a graph-structured map can be easily generated. The OSM data is structured using three basic entities: \textit{nodes}, \textit{ways} and \textit{relations}:
\begin{itemize}
    \item \textit{Nodes}: model point-shaped geometric elements described by their GPS coordinates and assigned unique node id. 
    \item \textit{Ways}: model line-shaped geometric objects that is formatted as two nodes and has a unique way id. 
    \item \textit{Relations}: represent relationships between nodes and ways. 
\end{itemize}  
According to such characteristics of data formation, we represent place and road as node and edge in the graph as shown in Fig. \ref{fig:osm_and_graph}. 

\begin{figure}[t]
    \centering
    \includegraphics[width=1\linewidth]{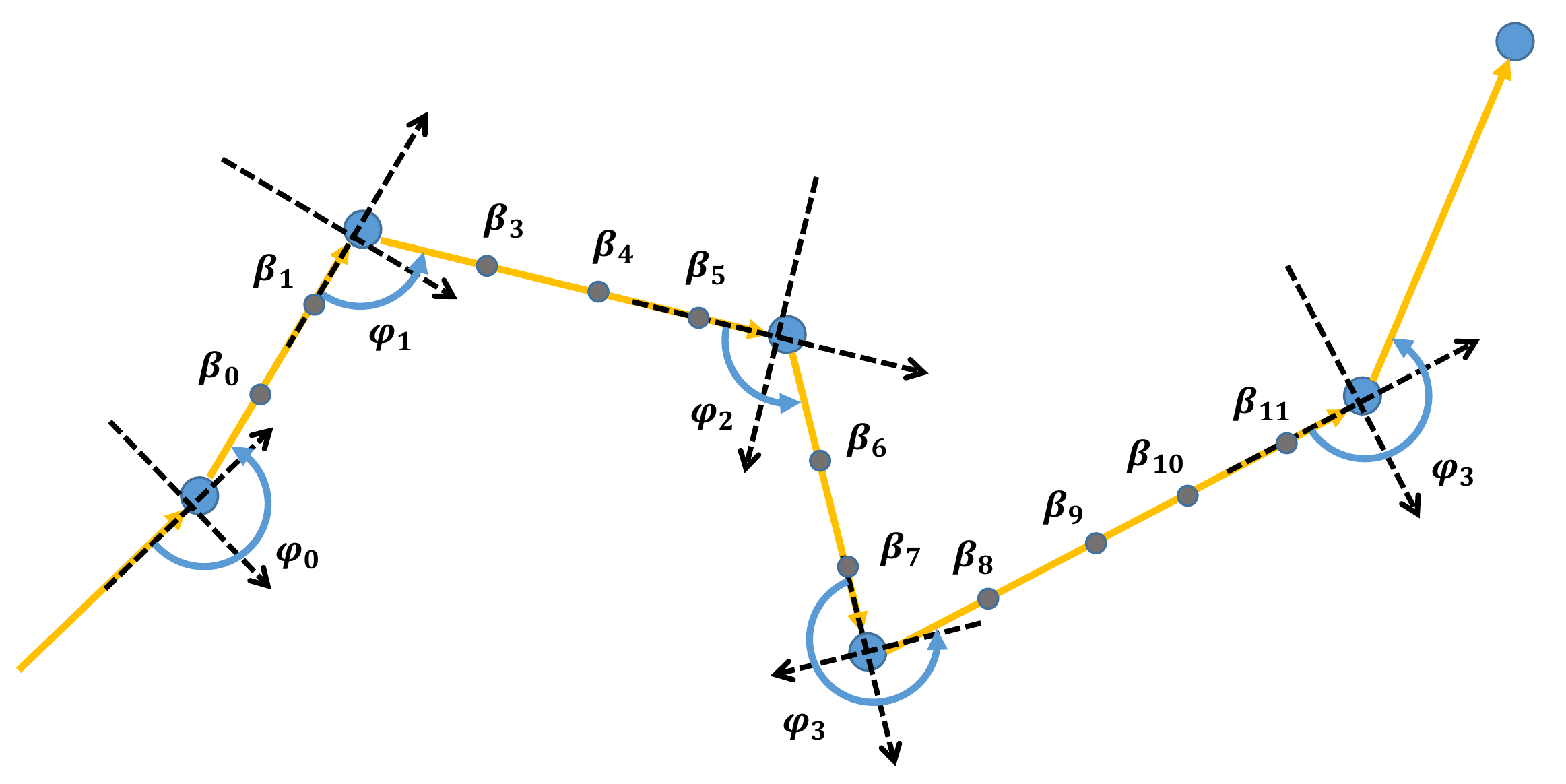}
    \caption{Motion trajectory representation using local turning angles and distance. The symbol $\varphi$ denotes real turning angle and $\beta$ indirectly represents distance information by inserting virtual nodes on road.}
    \label{fig:traj}
\end{figure}

The motion trajectory is generated as a result of object's motion on a topological map which consists of a sequence of distances traversed and turns taken. Hence, in this paper, we represent motion trajectory as a sequence of distances and turns which biological systems were shown to use for localization and navigation. Adopting the treatment of turning in a biological organism \cite{fyhn2004spatial}, \cite{hafting2005microstructure}, the turning angles are encoded in a local coordinate system centered at the intersection as shown in Fig. \ref{fig:traj}. In order to uniquely define the angles, the local coordinate system is generated at the entry direction of the intersection pointing in the turning direction. Then, the turning angle is calculated from previous, current and future nodes. For computational reasons, we quantized angles into 20 bins which provides our algorithm a discrete representation for turning angle. For the distance representation, we introduce virtual nodes in a uniform sample distance into a road segment between two nodes. Due to the the way they are generated, these virtual nodes introduce additional $180^\circ$ turning angle into the motion trajectory. Following this step, the distance information is transformed into an angle representation which is consistent with the turning angle representation. 

\begin{figure}[t]
    \centering
    \includegraphics[width=1\linewidth]{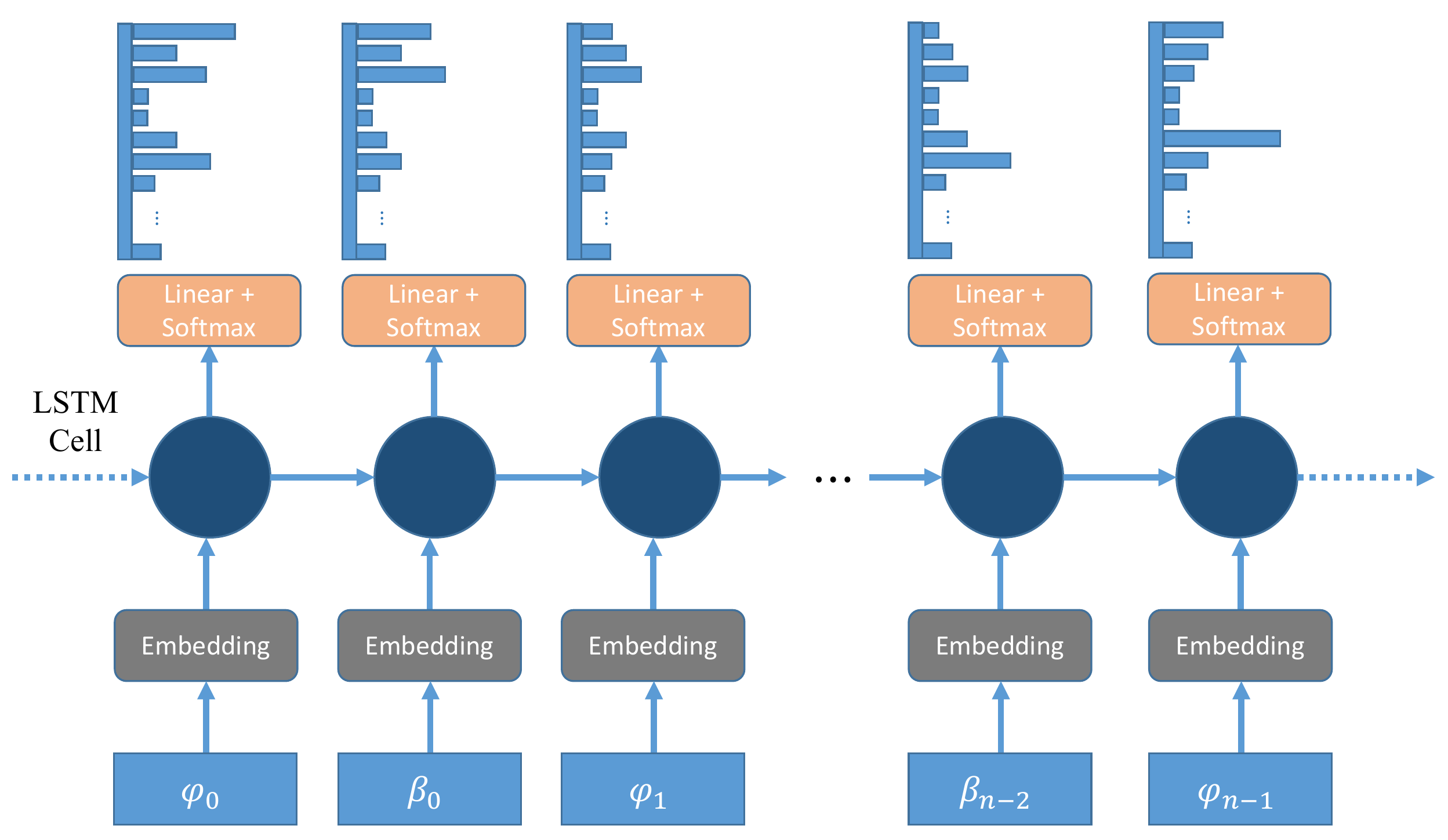}
    \caption{The recurrent neural network structure adopted for motion learning. Each input is a discrete angle following a embedding layer to transform scalar into high dimensional space. Each output is edge id on which object is located.}
    \label{fig:rnn}
\end{figure}

The trajectory learning strategy is realized by training an RNN \cite{elman1990finding}, which processes the motion trajectory as sequential data. In its general form, the RNN is the function of time $t$ that takes the current input $x_t$, the previous hidden state $h_{t-1}$ and produces a new hidden state $h_{t}$ through a non-linear activation function $f_\alpha$ and new output through a linear function $f_\beta$:
\begin{equation}
\label{equa:rnn}
    \begin{split}
        h_s &= f_\alpha(x_s, h_{s-1}) \quad \forall s=1, \dots, t   \\
        y_t &= f_\beta(h_t)  \\
    \end{split}
\end{equation}

\begin{figure*}[t]
\centering
\begin{subfigure}[t!]{0.46\linewidth}
  \includegraphics[width=1\linewidth]{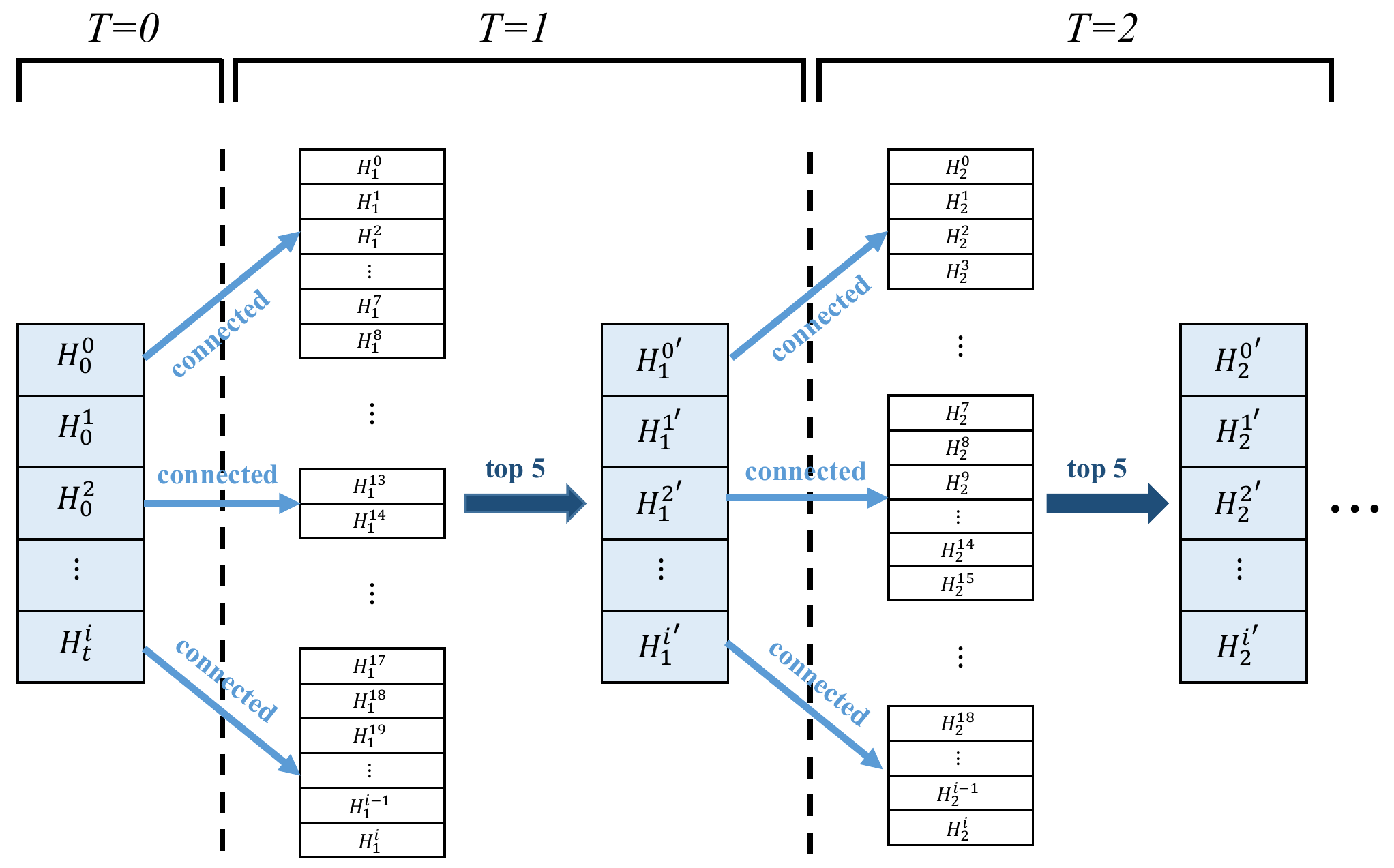}
  \caption{Hypotheses generation: strategy 1.}  
  \label{fig:mht_1} 
\end{subfigure} \hspace{2mm}
\begin{subfigure}[t!]{0.46\linewidth}
  \includegraphics[width=1\linewidth]{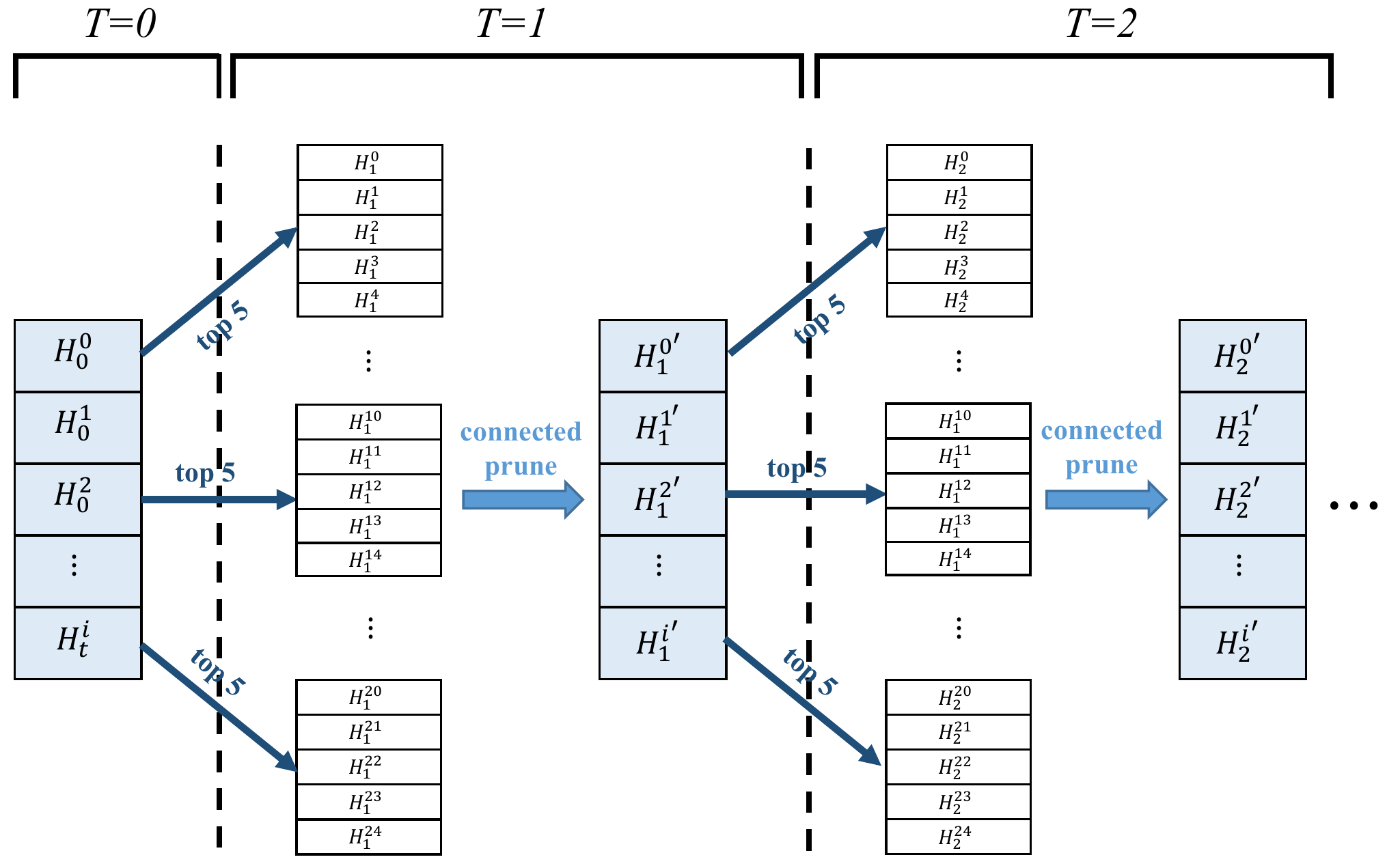}
  \caption{Hypotheses generation: strategy 2.}
  \label{fig:mht_2}
\end{subfigure} 
\caption{Two different hypotheses generation and elimination strategies. Track hypothesis at time $T  (T=1, 2. \dots, t)$ are shown by $H_t^{i}$ and $1<=i<=n$ denotes the the $i^{th}$ hypothesis from among $n$ hypotheses at time $t$. The light blue arrow denotes the edges connected to the edge denoting the current location hypothesis; the dark blue arrow refers to the selection of top $k$ hypotheses with highest probability at time $t$. The two strategies change in the way they select the top $k$ hypotheses.}
\label{fig:mht}
\end{figure*}
\begin{figure}[t]
    \centering
    \includegraphics[width=6.5cm, height=7cm]{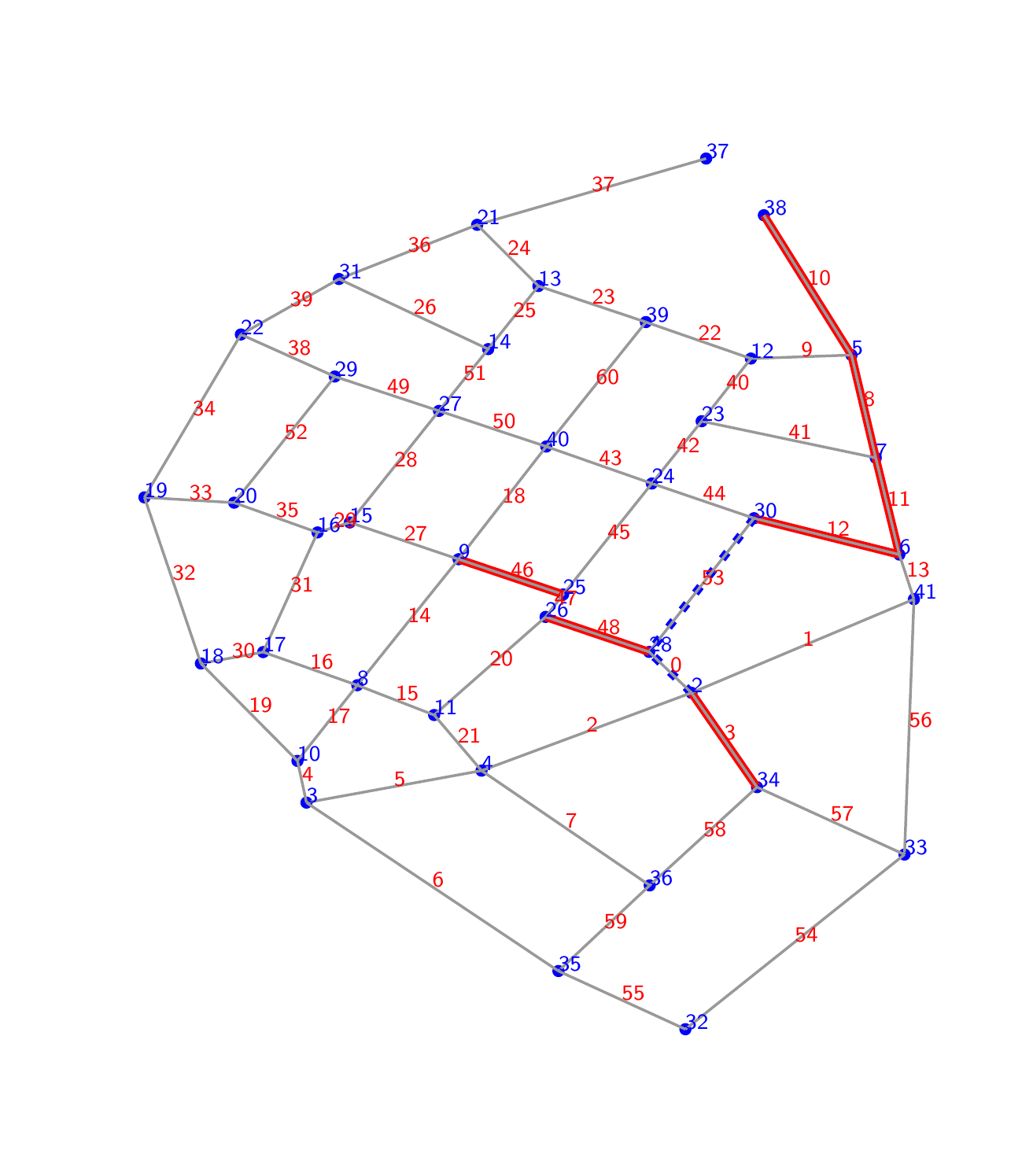}
    \caption{Inconsistent edge localization example. The red edge 46 and 48 are located in wrong place which should be located in blue dotted edge.}
    \label{fig:consistent}
\end{figure}

In our problem, each discrete turning angle in the sequence generated by the preprocessing phase is fed into the RNN one at a time as illustrated in Fig. \ref{fig:rnn}. An embedding layer is attached to the input layer as a high-dimensional representation of a discrete scalar input. The output at each time step generates the edge on which the object makes its last turn. 
Finally, we use a \textit{softmax} function: 
\begin{equation}
\label{equa:softmax}
    P(Y=i|y) = softmax(y)=\frac{e^{y}}{\sum_{j=0}^{k}e^{y}}
\end{equation} 
to calculate the probability of each output edge class which corresponds to unique edge id in a given map graph, where $y$ is the linear output in each time, $Y=i$ denotes the edge id which is equal to $i$. This equation shows the probability of the each edge output in each time as shown in Fig. \ref{fig:rnn}. To train the network, we use negative log likelihood loss (NLL) 
on the edge probabilities:
\begin{equation}
    L_i = -log(p_{y_i})
\end{equation}

\subsection{Temporal Consistency}
The trajectory learning process takes a sequence of motion features as its input and its corresponding edge ids as output to train the underlying RNN. The training is performed using edge id sequence which implicitly seems to introduce temporal consistence to the localized positions. However, the output depends only on the past state which breaks the temporal consistency and that the sequence of output edge ids become inconsistent on the map. An example of this undesired phenomena is shown in Fig. \ref{fig:consistent} where the algorithm skips the blue dotted line which is what should have be predicted and instead chooses two red edges that are not connected hence are temporally inconsistent. 
In order to deal with  temporal inconsistencies, we propose two different  multiple hypotheses generation and elimination strategies as a part of the  incremental localization illustrated in Fig. \ref{fig:mht}. We define a hypothesis as the object being on an edge between two nodes (virtual and real). Assuming initial position is unknown, an object can be on any edge which results in initially as many hypotheses as there are edges (real and virtual) on the map. As the object moves, each hypothesis at time $t-1$, $\{H_{t-1}^{i}\}$ generates new hypotheses at time $t$, $\{H_{t}^{ij}\}=(\{H_{t-1}^{i}\}, \{H_{t}^{j}\})$ based on the object moving from the current edge to any connecting edge. The likelihood of the new hypotheses is defined by the likelihood ratio as:
\begin{equation}
    P(H_t^{ij}|H_{t-1}^{i}) = \frac{P(H_t^{ij}, H_{t-1}^{i})}{P(H_{t-1}^{i})}
\end{equation}
where, $i = 1, \dots, n_1, \enskip j = 1, \dots, n_2$ and $n_1, n_2$, are the number of hypotheses in previous time and current time, respectively. $P(H_t^{ij}|H_{t-1}^{i})$ is the conditional probability of hypothesis at current time instant. Below we introduce two strategies for hypotheses generation. 

{\bf Strategy 1:} As shown in Fig. \ref{fig:mht_1}, the algorithm generates a new location hypothesis at current time $t$ for each edge connected to the previous edge defined by the hypothesis at time $t-1$. At $t=0$, all edges in the map are used to generate location hypotheses as we do not know where the object starts the traverse. In subsequent time instants, the probability of each of the hypotheses is computed by conditioning it on the previous location hypothesis. The number of hypotheses in this approach can be combinatorially explosive, hence, for computational reasons, one can keep only the top $k$ location hypotheses. In our experiments we set $k=5$. Considering the way each hypothesis is generated based on the connectivity rule, temporal consistency of the location estimation is guaranteed. 

{\bf Strategy 2:} As shown in Fig. \ref{fig:mht_2}, the algorithm generates $k$ location hypotheses with the highest probability based on the previous hypothesis at time $t-1$. The strategy proceeds by  eliminating improbable hypotheses based on the edge connectivity constraint. While this method produces temporally consistent trajectories in the form of sequence of edges, it suffers from the the possibility that the true prediction might not be in the first top $k$ hypotheses which will lead to an inconsistent prediction. Later in the experiments section, we will show results that illustrates such inconsistencies and compare them with the strategy 1. 

\subsection{Motion Trajectory Generation}
\begin{figure}[t]
    \centering
    \includegraphics[width=0.9\linewidth]{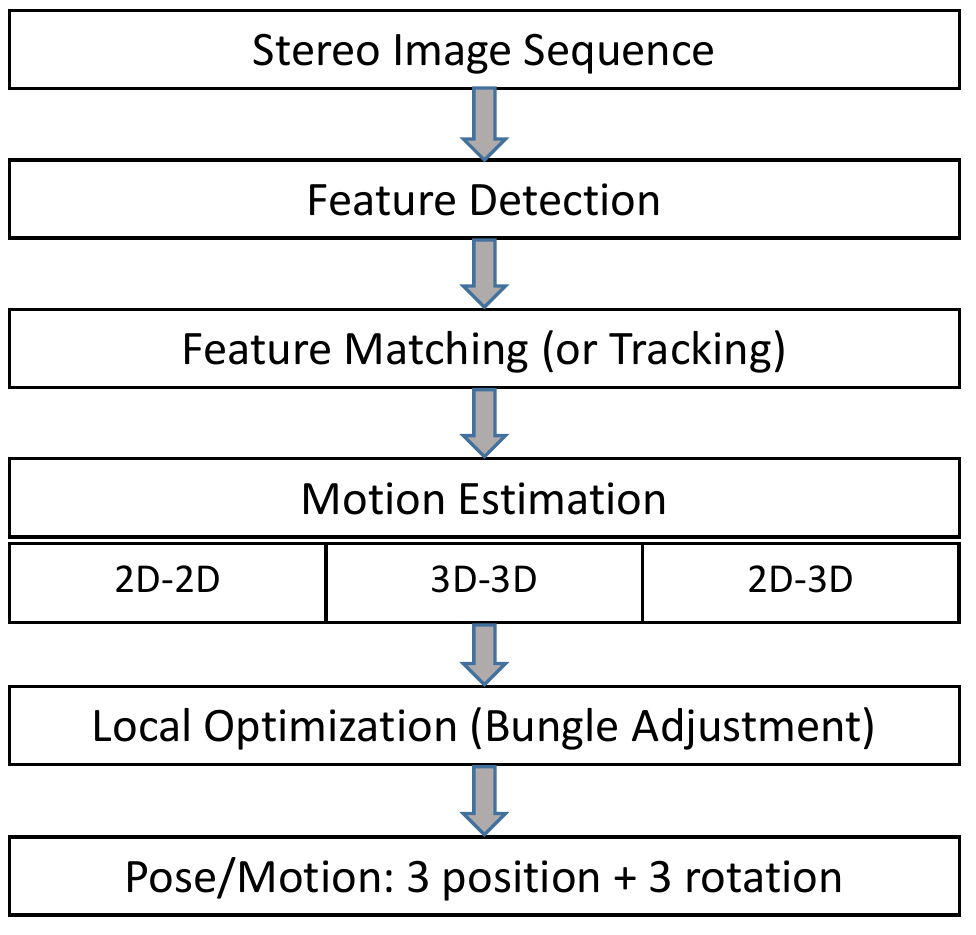}
    \caption{Stereo Visual Odometry Pipeline (Adapted from \cite{scaramuzza2011visual}).}
    \label{fig:vo}
\end{figure}
The distance information used as a motion features is encoded with the object's trajectory which requires us to generate a metric trajectory by  utilizing the stereo visual odometry.

Visual odometry estimates a motion trajectory with six degrees of freedom including 3D position and orientation of the platform using image data. Visual odometry can  estimate the relative motion trajectory from a single camera (monocular) setup or a stereo camera setup. In this paper, we use the stereo setup with known calibration.
Using this information, we estimate the relative rotation $\mathcal{R}$ and translation $\mathcal{T}$ by detecting and matching features between consecutive stereo frames. These matching points are then used in photogrammetric intersection framework to estimate instantaneous relative motion $\mathcal{R}$ and $\mathcal{T}$. Then, we are able to obtain a sequence of metric points as motion trajectory. We display the pipeline we have used  for estimating motion trajectory in Fig. \ref{fig:vo}. The details of related techniques can refer to \cite{hartley2003multiple}.

\section{Experiments and Results}
\label{sec:experi}
In this section, we first introduce the dataset used in this paper including data generation by graph search algorithm for training and real data generated by stereo visual odometry for testing. Then, the details of learning process is presented. Next, two different results under raw and consistent case are exhibited. 

\subsection{Training data generation for trajectory learning}

\begin{figure}[t]
\centering
\begin{subfigure}[h]{0.24\textwidth}
   \includegraphics[width=4.2cm, height=4cm]{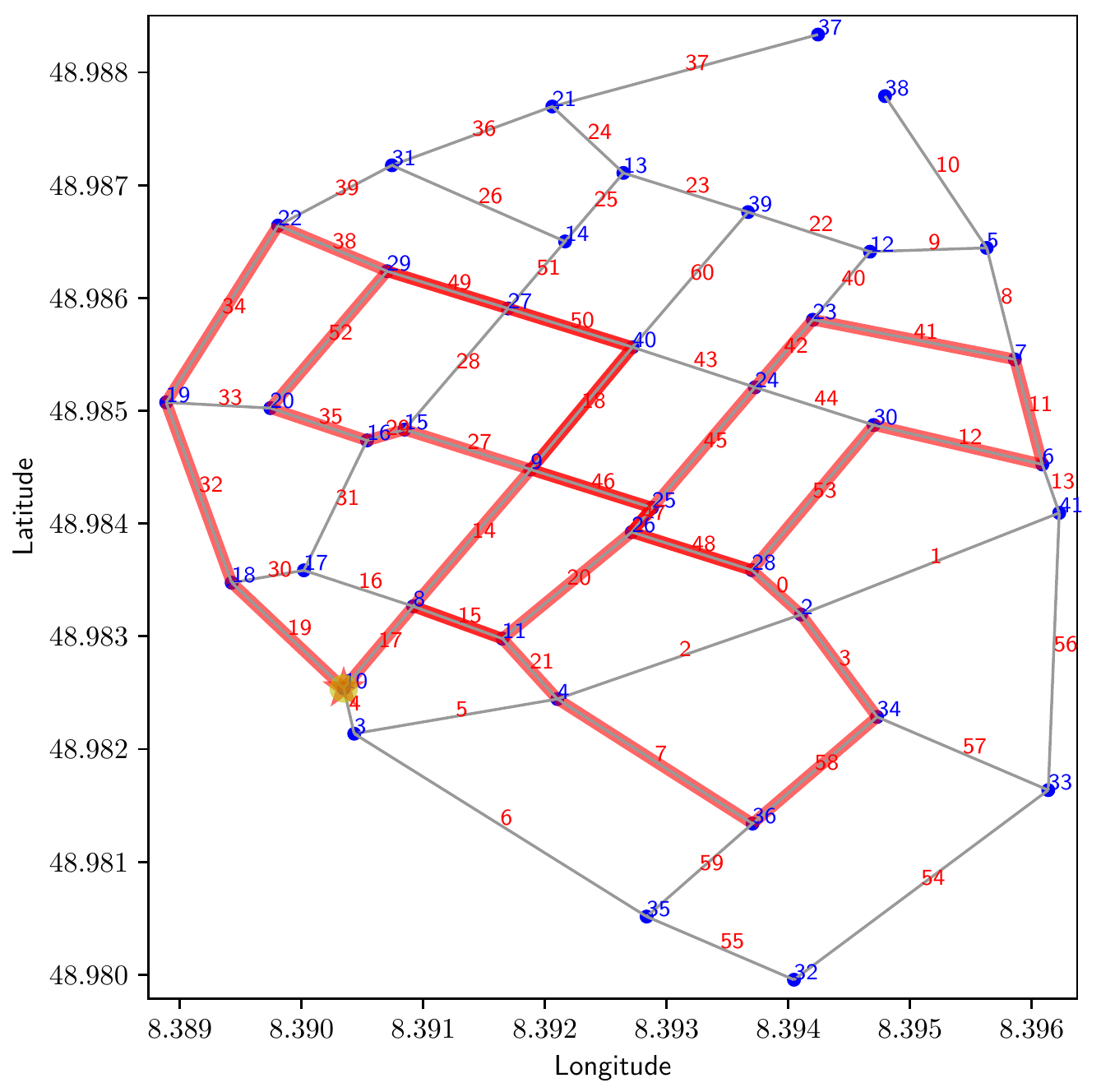}
   \caption{Training trajectory example.}
   \label{fig:data_syn} 
\end{subfigure}
\begin{subfigure}[h]{0.24\textwidth}
   \includegraphics[width=4.2cm, height=4cm]{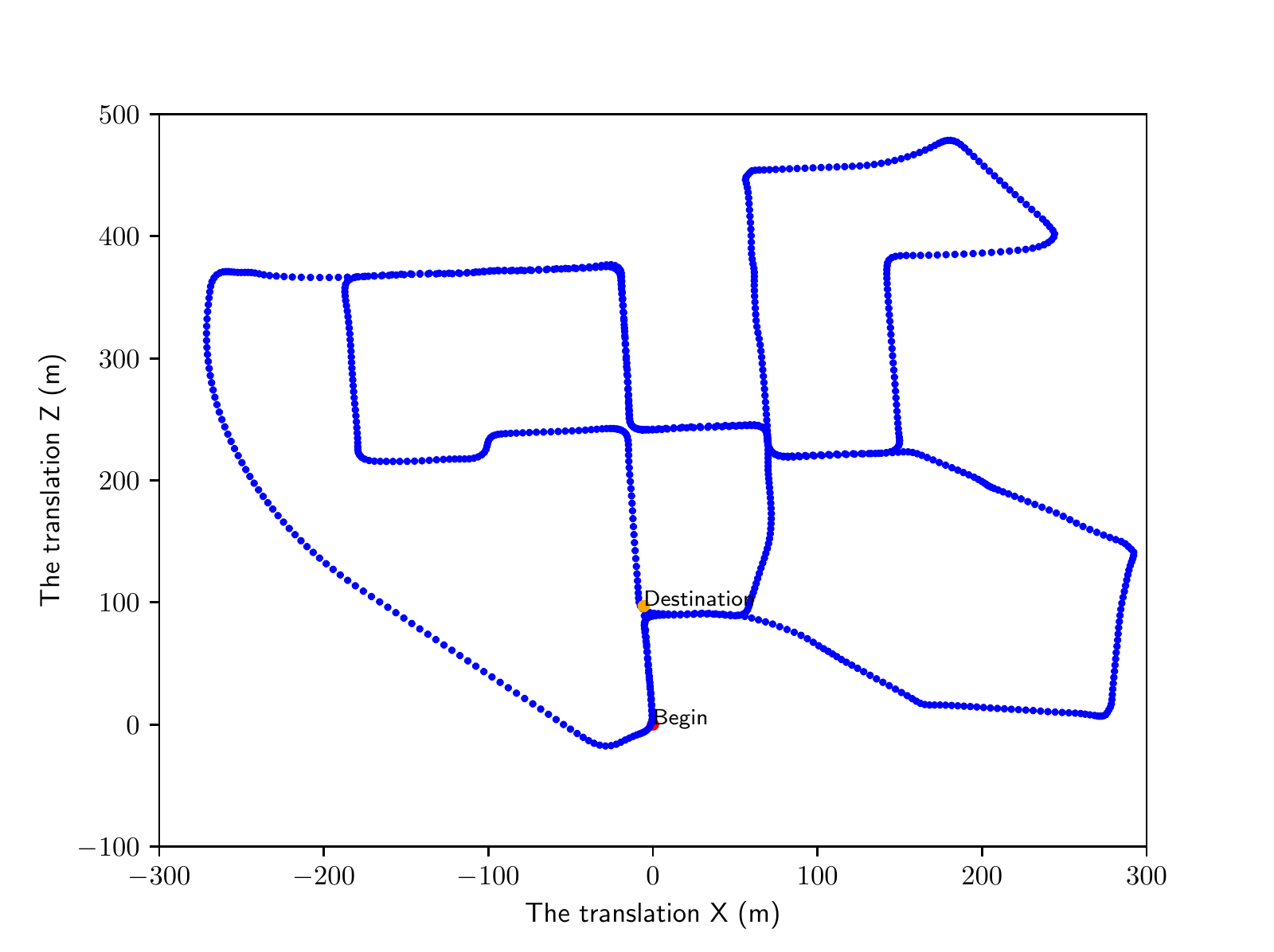}
   \caption{Test trajectory example.}
   \label{fig:data_re}
\end{subfigure}
\caption{Training dataset generated by graph path search, and test dataset generated by stereo visual odometry.}
\label{fig:data_example}
\end{figure}

\begin{table}[t]
    \centering
    \caption{The original dataset statistics.}
    \label{tab:data} 
    \begin{tabular}{l c} 
        \toprule 
        \textbf{Data} &  \textbf{Statistics} \\
        \midrule 
        Topological Map  &  40 nodes, 61 edges  \\ 
        Trajectory Length  &  10 nodes  \\ 
        All Trajectories  &  17537  \\ 
        All Classes  & 61  \\ 
        Input Feature Space  & 20  \\ 
        Training Trajectories  & 17536  \\ 
        Training Output Classes  & 61  \\ 
         \bottomrule 
    \end{tabular} 
\end{table}

The proposed trajectory learning process uses the motion trajectories and topological map. For training purpose, it is necessary to generate a large number of motion trajectories using the given map. In theory, one select any source node on the map and can generate all possible trajectories from that node to any other target node with different lengths. This treatment, however, will generate an unnecessarily large number of overlapping trajectories. Instead, we select fixed number of nodes on a given path and use a modified depth-first search algorithm \cite{robert2002algorithms} to generate the training trajectories. Specifically, given a source node and a target node, depth-first search produces all paths where the vertices and edges are distinct. In our implementation, we choose 10 nodes to generate the training trajectories which allow adequate distance and turning information to use and avoid unnecessarily long sequences causing overfitting problems in training step. An example training trajectory is shown in  Fig. \ref{fig:data_syn}. After the generation of the training trajectories, we introduce the virtual nodes illustrated in Fig. \ref{fig:traj} that simulates the grid cells in hippocampus. This modified trajectory is then used to produce the angle sequence as training data input to RNN. The output data that corresponds to this training data are the edge ids on which the last turn occurs. The dataset statistics is shown in the Table \ref{tab:data}. 

\subsection{Real trajectory data used for testing}  

For testing our approach, we use the publicly available KITTI dataset \cite{geiger2012we}. The benchmark dataset consists of stereo image sequences collected from different areas of which part of sequences provide the ground-truth GPS information. In this paper, we select the sequence (id=00) containing 4541 stereo image pairs which  provides a complete trajectory of a vehicle traveling around Karlsruhe, Germany as shown in Fig. \ref{fig:data_re}. Our test data is subsets of this complete trajectory. We used the VINS-Fusion \cite{qin2019general} as the visual odometry approach to generate  3D trajectories. 

\subsection{Trajectory Learning Setup}

We formulate the  motion learning as a conditional variable sequence prediction problem using RNN as shown in Fig. \ref{fig:rnn}. The above generated training trajectory dataset is used to train our recurrent neural network where each input is discrete angle and output is corresponding edge id. Considering that learning map is a multi-class classification problem, at each time instant, the cross-entropy function is utilized as the loss function for measuring the training performance. We choose the stochastic gradient descent as optimization method. 

\begin{figure}[t]
\centering
\begin{subfigure}[h]{0.24\textwidth}
   \includegraphics[width=4.3cm, height=3.3cm]{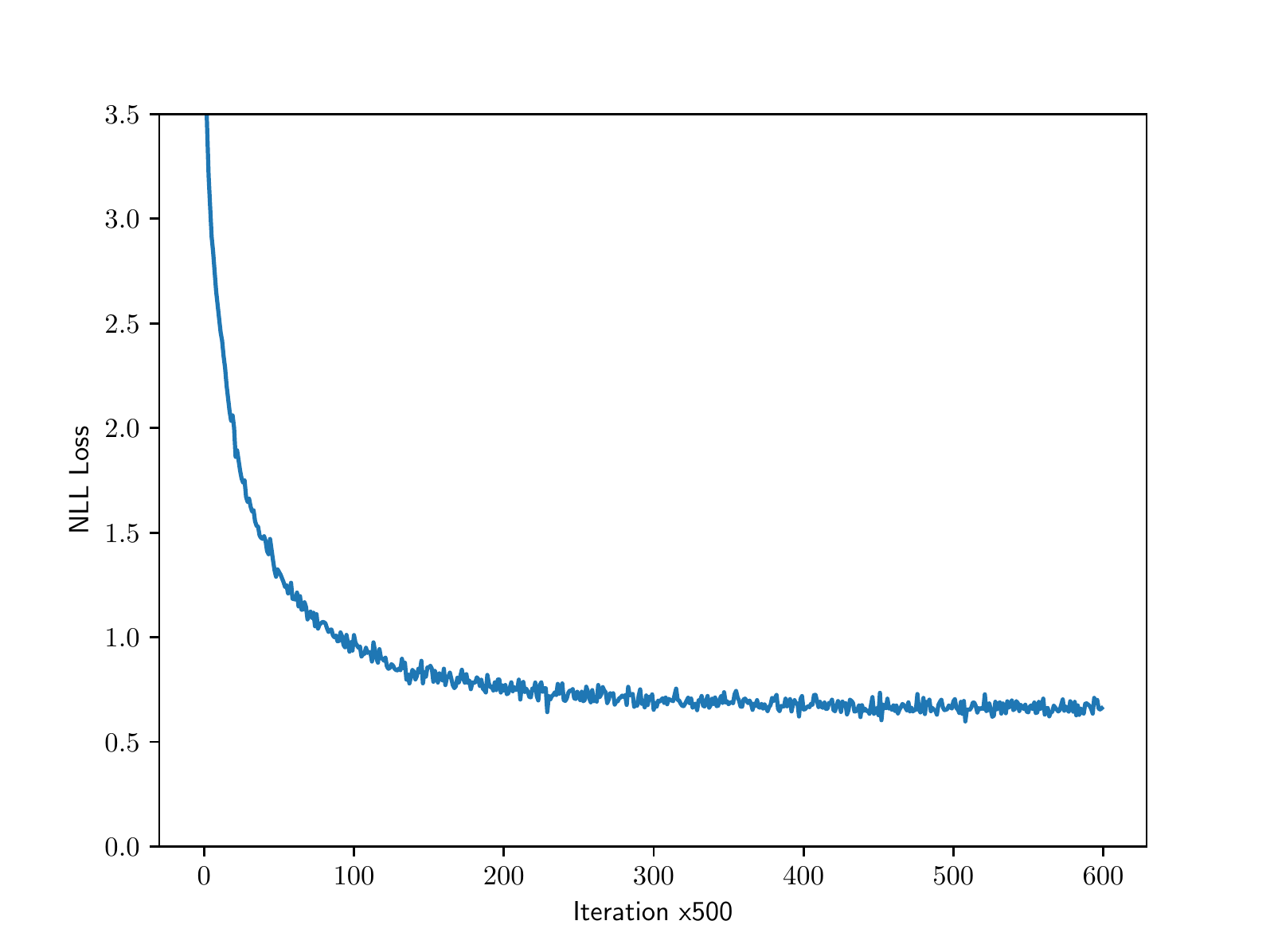}
   \caption{Training loss.}
   \label{fig:train_loss} 
\end{subfigure}
\begin{subfigure}[h]{0.24\textwidth}
   \includegraphics[width=4.3cm, height=3.3cm]{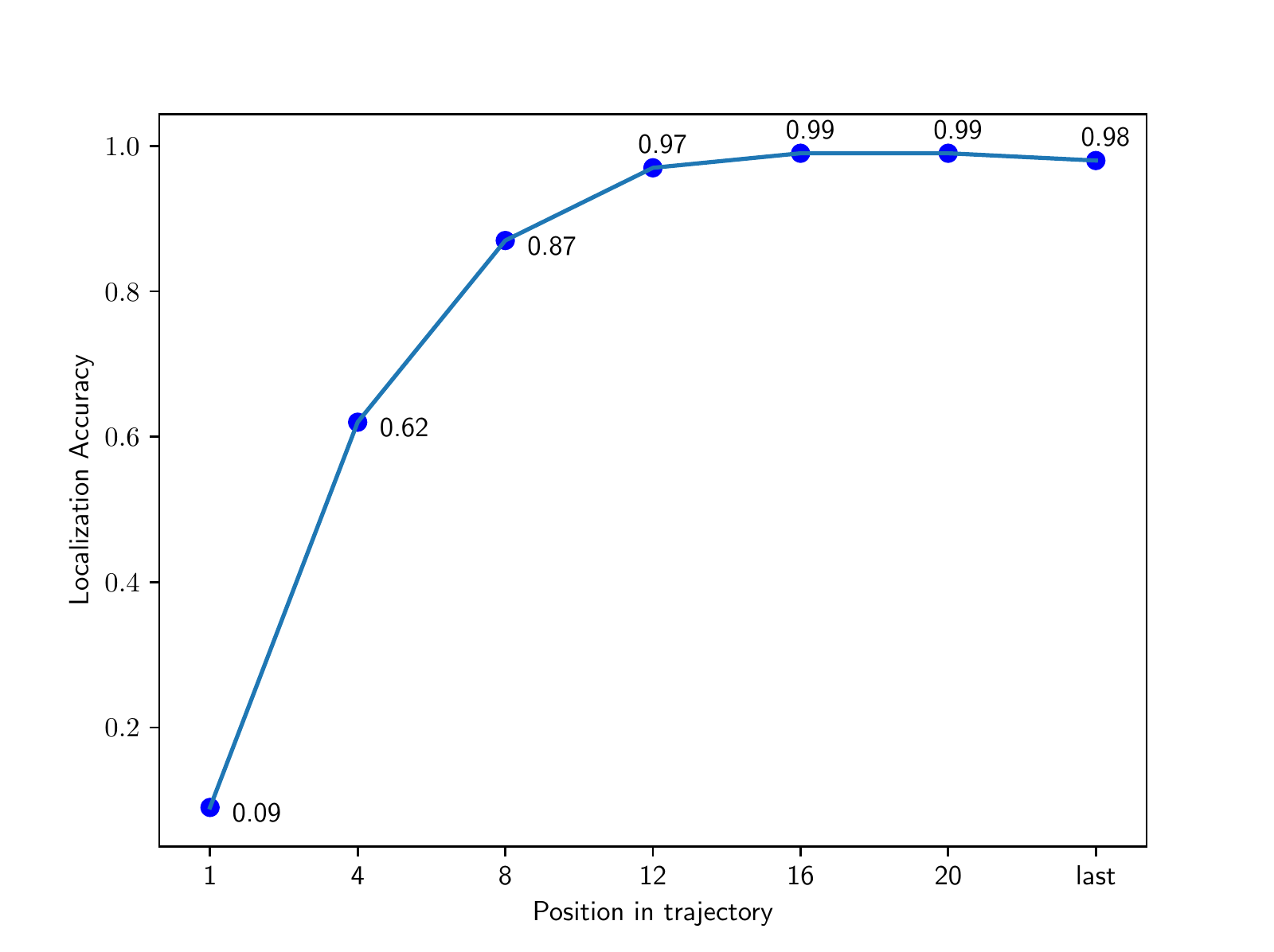}
   \caption{Accuracy.}
   \label{fig:acc}
\end{subfigure}
\caption{Change of training loss and final localization accuracy with different length of trajectory where $x$ axis denote $i$-th position in trajectory. Due to the variable length of trajectory, the label ``last'' only indicates the last position in trajectory }
\label{fig:loss_acc}
\end{figure}

For the hyperparameter settings, we only use one-layer RNN with 128 hidden units. The initial weights are all set to zero and learning rate is set to 0.01. We use batch size of one trajectory and the total number of iterations is set to 6 times the size of training data. At each iteration that batch is generated randomly. From Fig. \ref{fig:train_loss}, we can see that the loss is reducing to zero with more epochs which demonstrates convergence suggesting that the neural network model has learned the map. 

In Fig. \ref{fig:acc}, we show the accuracy of localization as a function of length of trajectory segment. The accuracy is computed as: 
\begin{equation}
\label{equ:acc}
    Accuracy(i) = \frac{1}{N}\sum_{j=0}^N T_{ji}
\end{equation}
where, $T_{ji}$ is correctness of prediction, $\{0,1\}$, on the $i$-th node of trajectory $j$ and $N$ is the total number of trajectories used to generate the plot. In this plot,  $N$ is total number of training trajectories. We should note that using the accuracy at the $i$-th node refers to the accuracy of having a trajectory of length $i$. The $x$-axis in Fig. \ref{fig:acc} represents the node used to compute the position accuracy at that node. 
The node axis includes the inserted ``virtual'' nodes, which leads to differences in trajectory lengths for trajectories with the same number of map nodes. Therefore, the use ``last'' phrase on the $x$-axis refers to the last node of each of the $N$ trajectories. 
From the Fig. \ref{fig:acc}, it can be observed that the accuracy increases with increase in the length of trajectory from 1 to entire trajectory segment. Considering the input angle and output edge are mapped to each other, as vehicle traverses the map, the spatial uncertainty of its whereabouts reduces to a single location hypothesis. This observation is supported by the fact that the longer the trajectory becomes, it would fit only to a certain map location due to increase in accuracy. 

\subsection{Consistent Geolocalization}

\begin{figure*}[t!]
\centering
\begin{tabular}{p{0.15\linewidth} p{0.15\linewidth} p{0.15\linewidth} p{0.15\linewidth} p{0.15\linewidth} p{0.15\linewidth}}
    Motion trajectory & Ground truth & Initial hypotheses & Inconsistent  & Consistent by  & Consistent by \\  
    & & & prediction & strategy 1 &  strategy 2  \\
\end{tabular}  \\
\vspace{0.4em}
\begin{subfigure}[t!]{0.15\linewidth}
  \includegraphics[width=2.5cm, height=3.5cm]{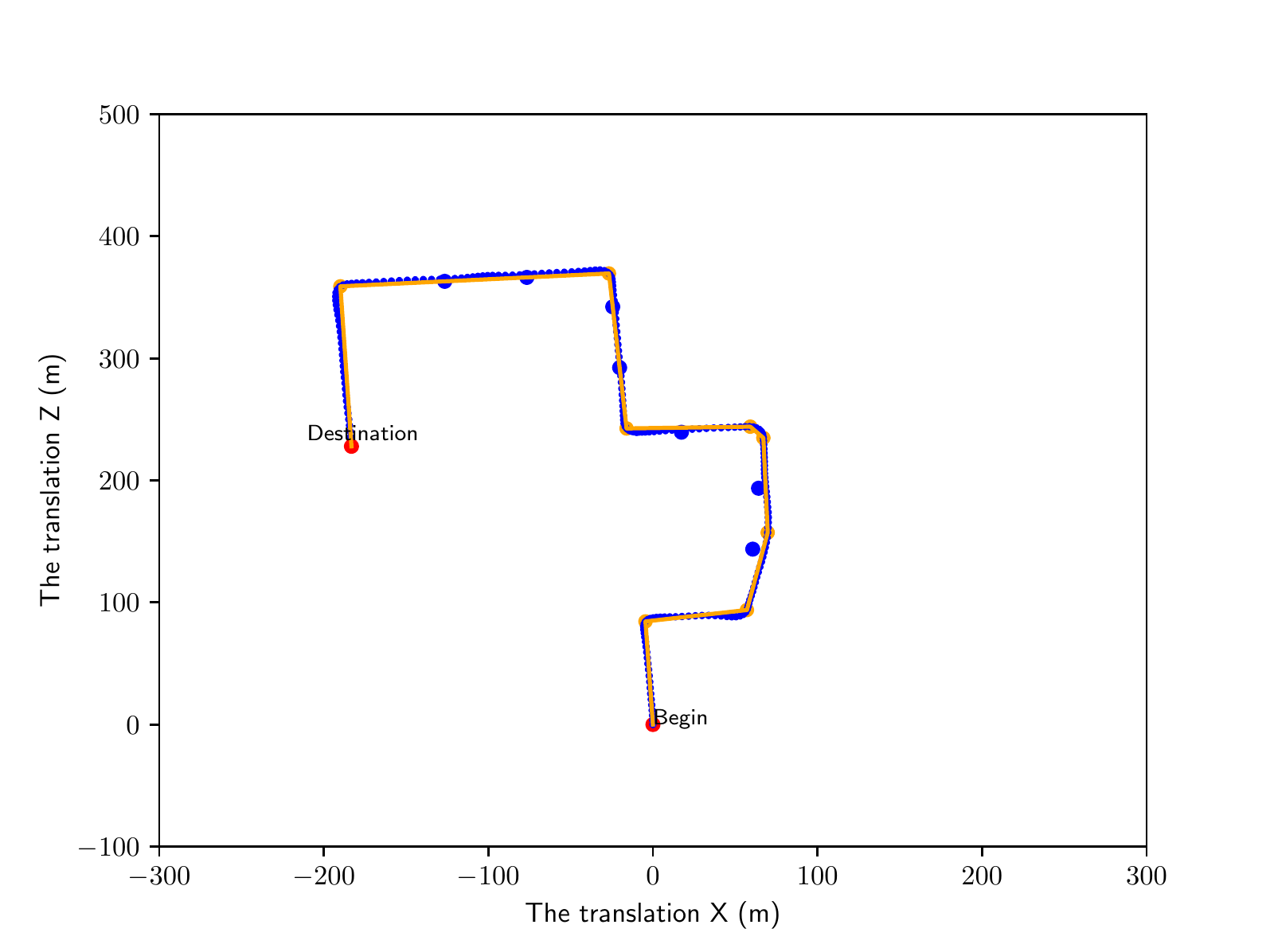}
\end{subfigure}  \hspace{0.5mm}
\begin{subfigure}[t!]{0.15\linewidth}
  \includegraphics[width=2.5cm, height=3.5cm]{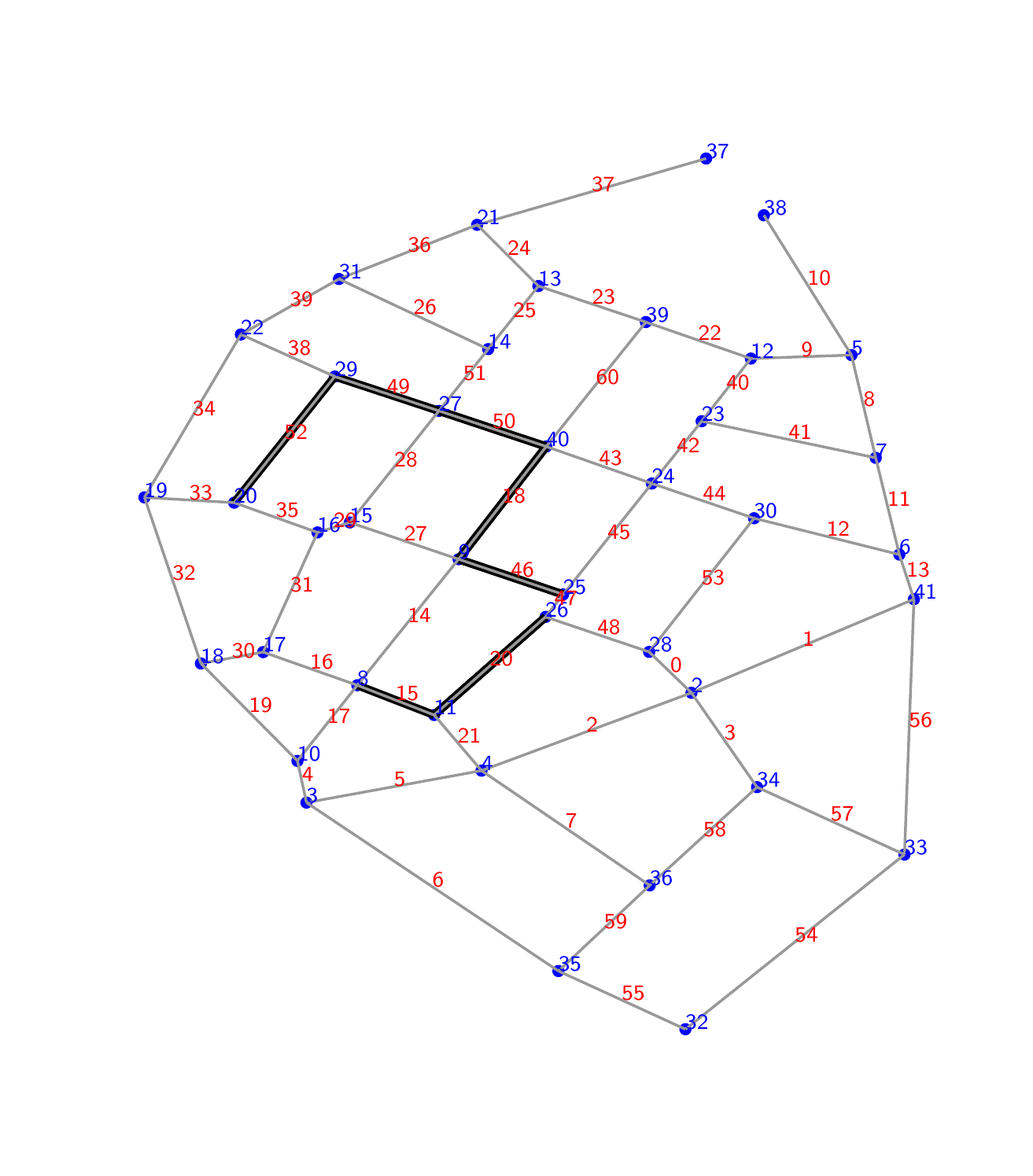}
\end{subfigure}  \hspace{0.5mm}
\begin{subfigure}[t!]{0.15\linewidth}
  \includegraphics[width=2.5cm, height=3.5cm]{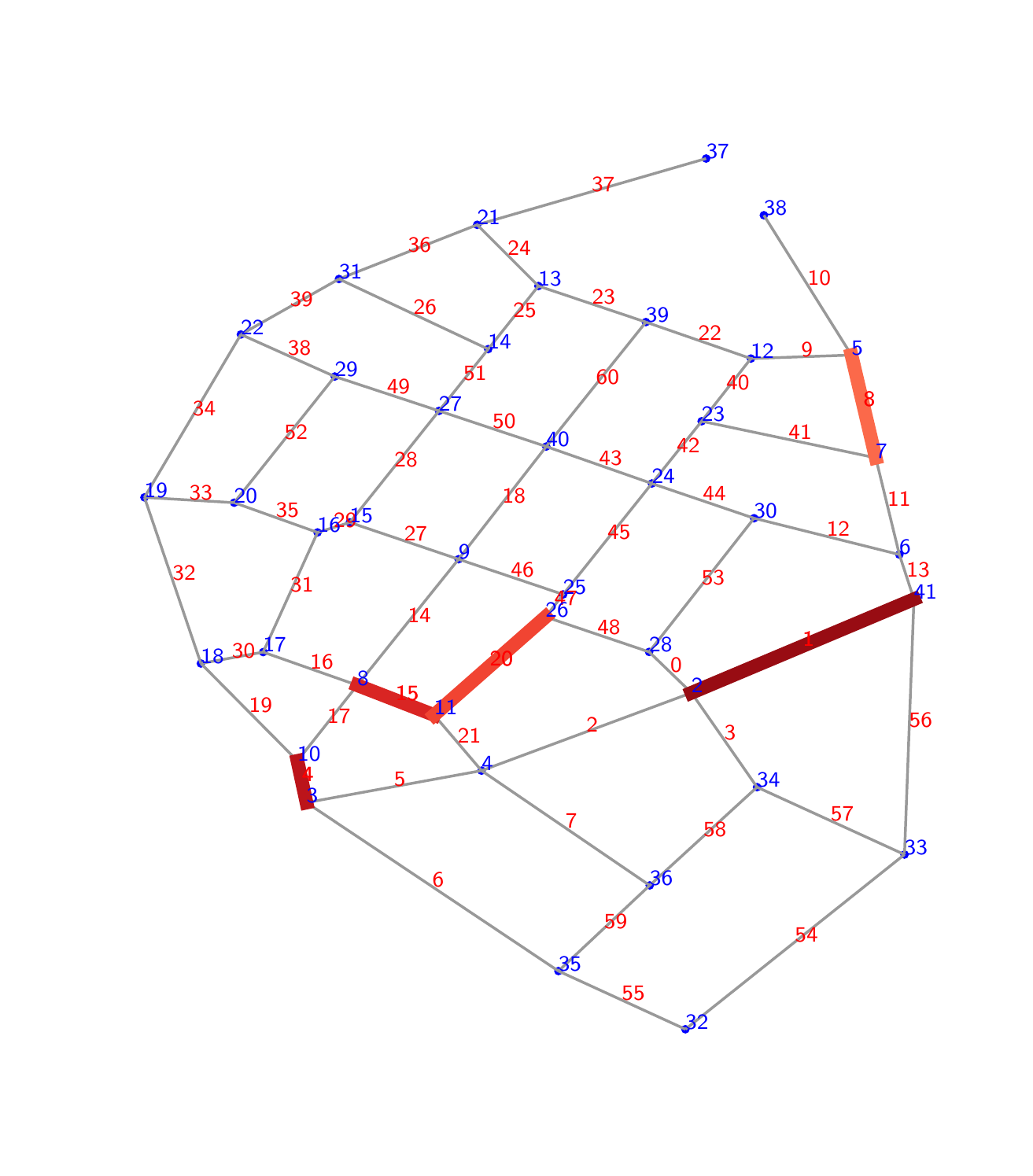}
\end{subfigure}   \hspace{0.5mm}
\begin{subfigure}[t!]{0.15\linewidth}
  \includegraphics[width=2.5cm, height=3.5cm]{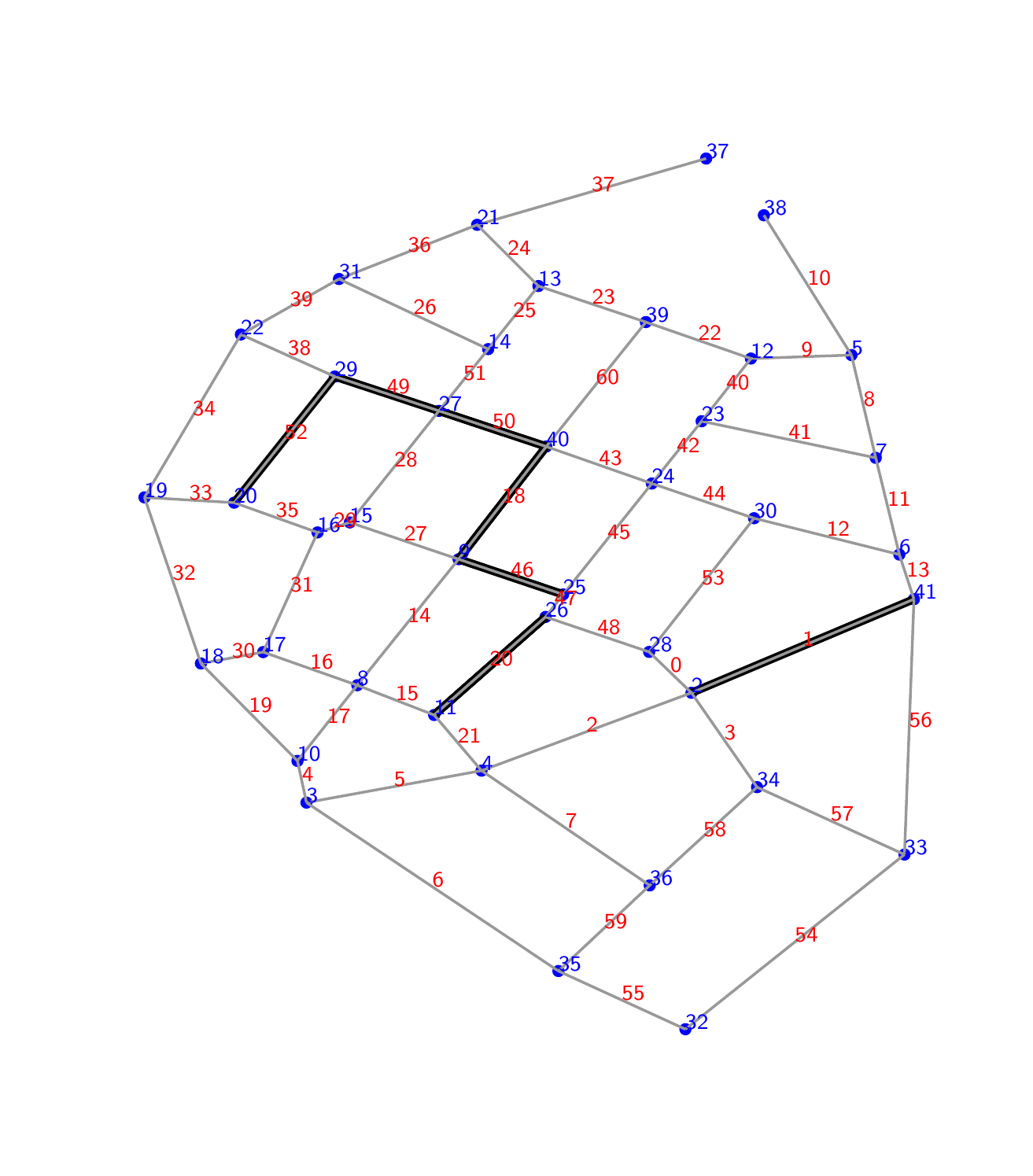}
\end{subfigure}   \hspace{0.5mm}
\begin{subfigure}[t!]{0.15\linewidth}
  \includegraphics[width=2.5cm, height=3.5cm]{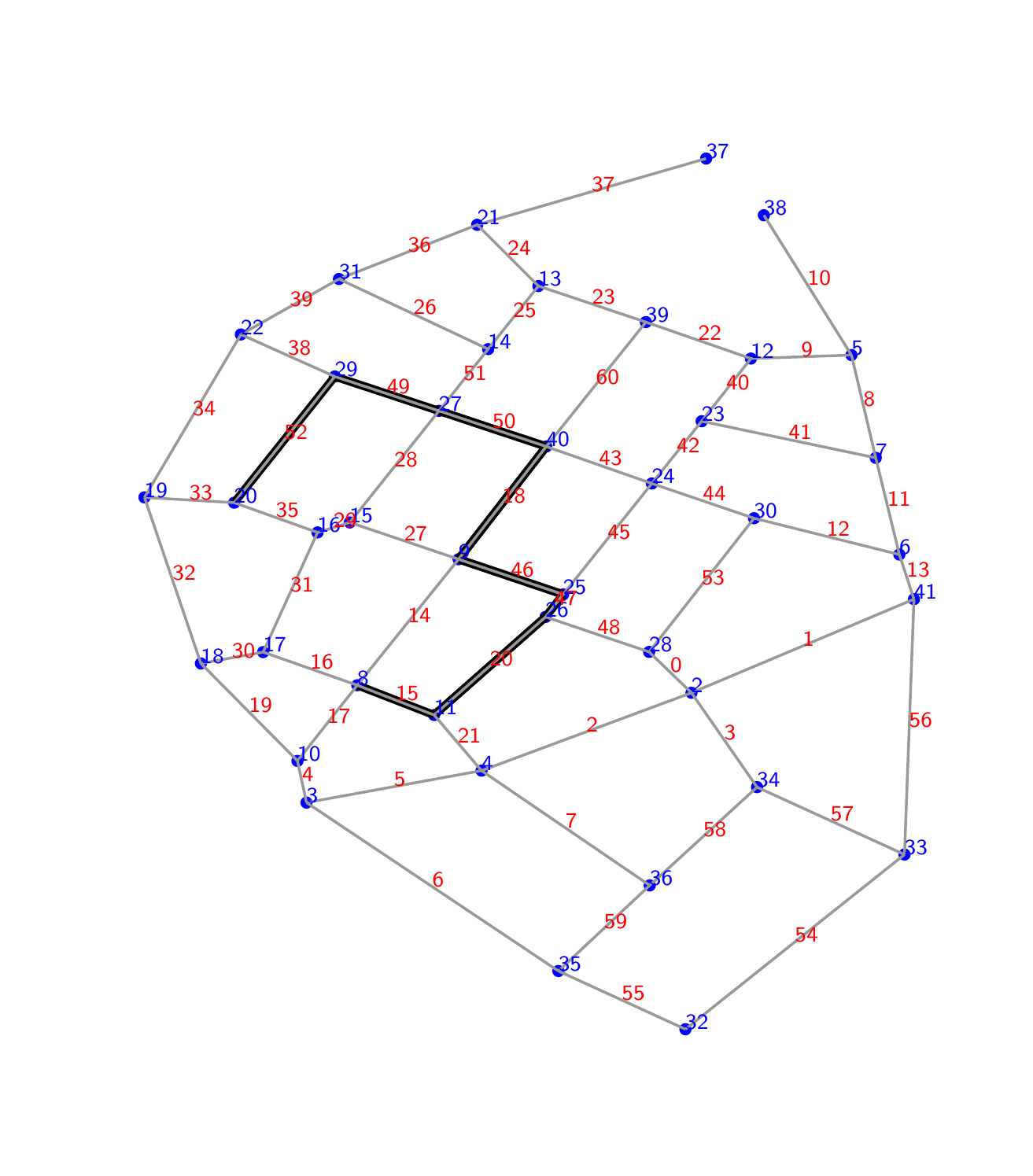}
\end{subfigure}   \hspace{0.5mm}
\begin{subfigure}[t!]{0.15\linewidth}
  \includegraphics[width=2.5cm, height=3.5cm]{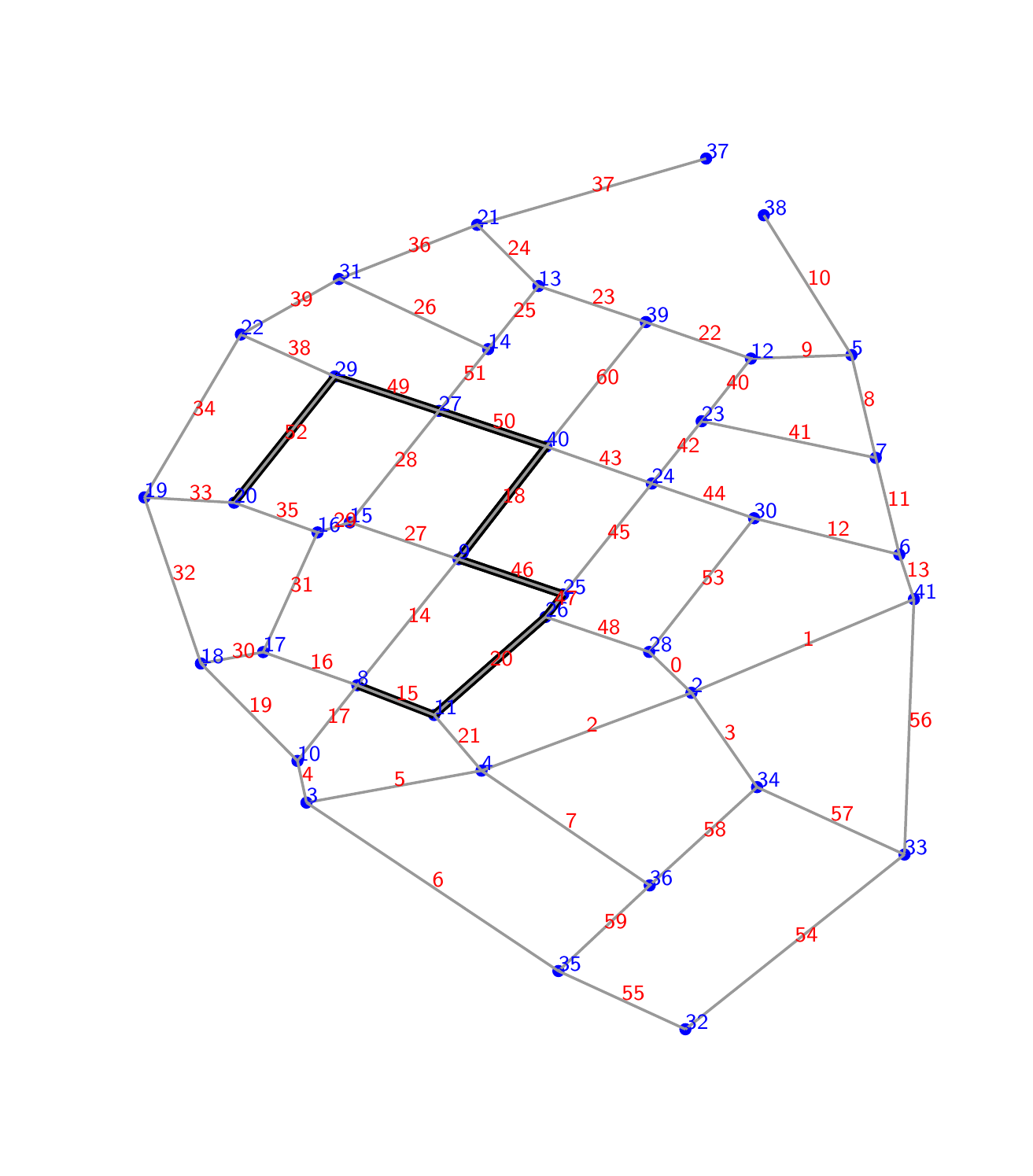}
\end{subfigure} 

\vspace{2mm}
\begin{subfigure}[t!]{0.15\linewidth}
  \includegraphics[width=2.5cm, height=3.5cm]{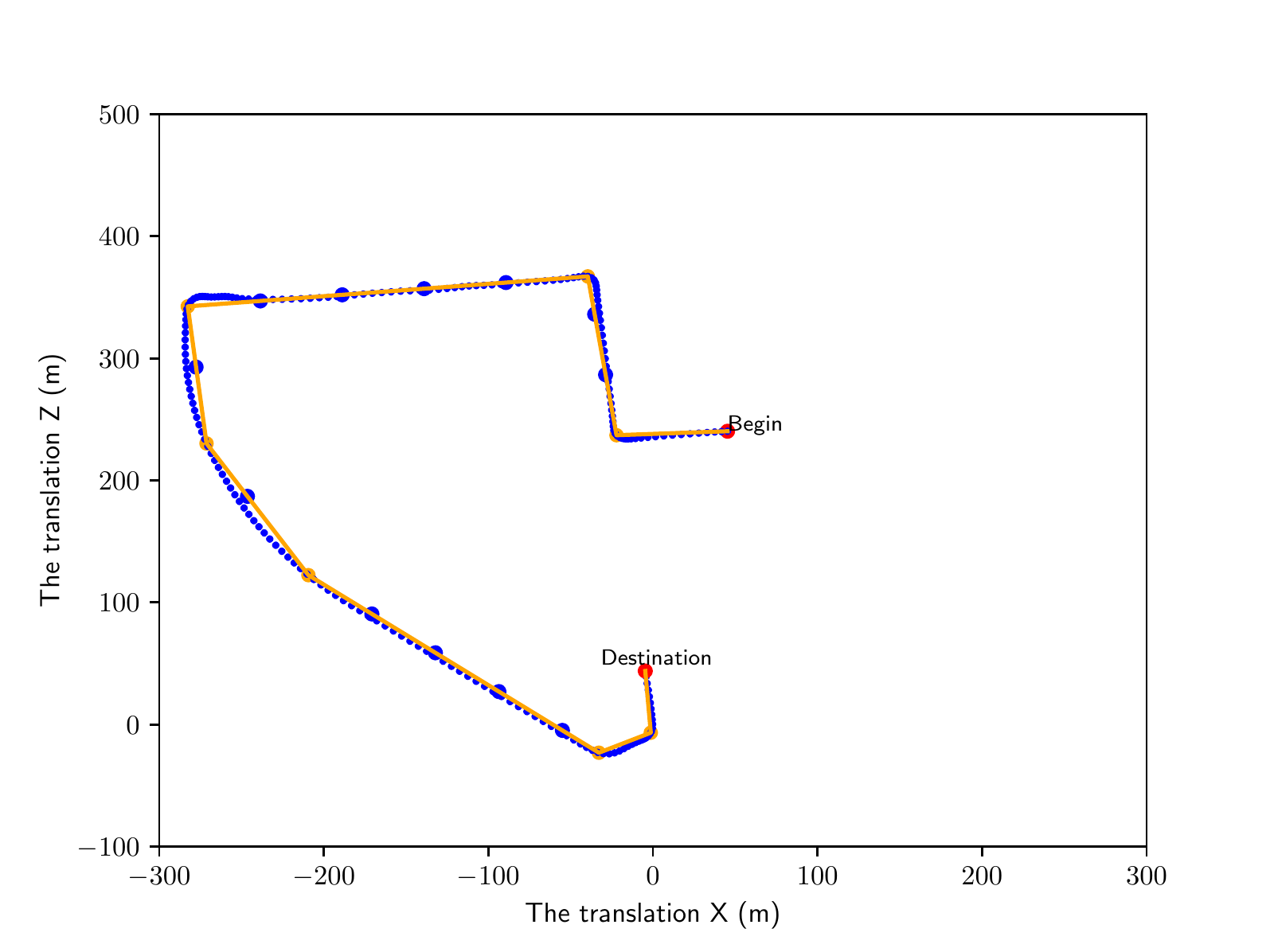}
\end{subfigure}  \hspace{0.5mm}
\begin{subfigure}[t!]{0.15\linewidth}
  \includegraphics[width=2.5cm, height=3.5cm]{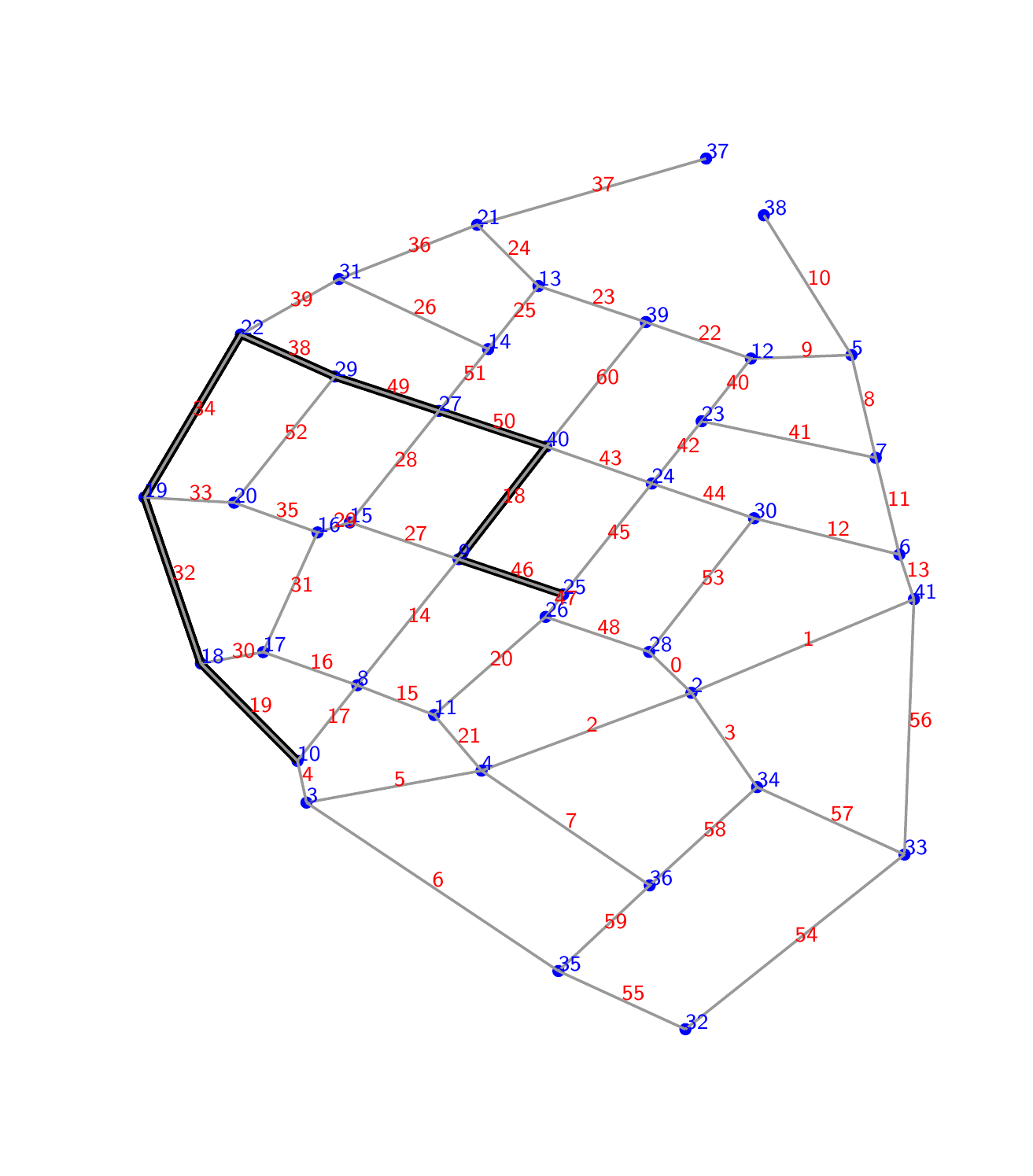}
\end{subfigure}   \hspace{0.5mm}
\begin{subfigure}[t!]{0.15\linewidth}
  \includegraphics[width=2.5cm, height=3.5cm]{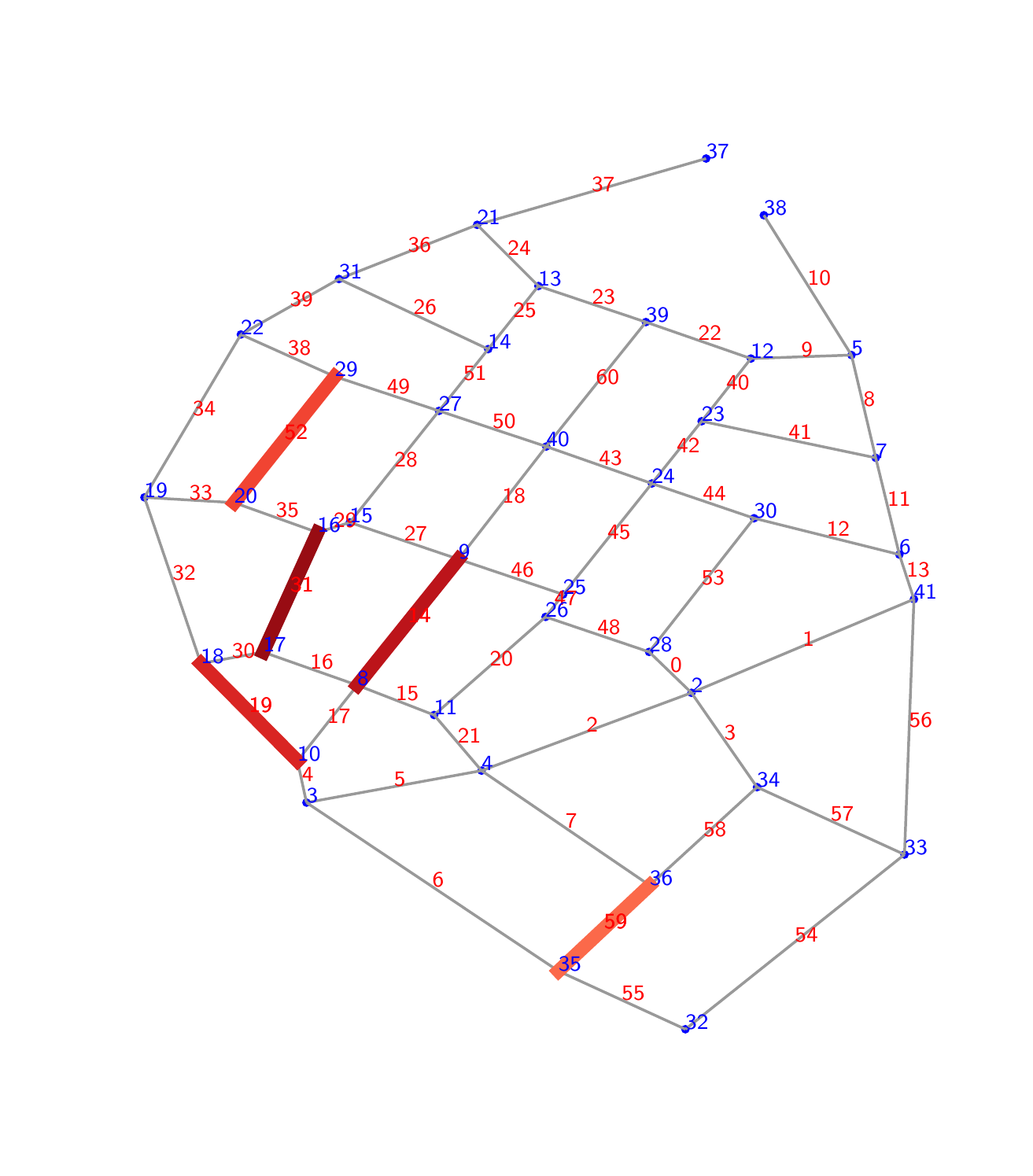}
\end{subfigure}   \hspace{0.5mm}
\begin{subfigure}[t!]{0.15\linewidth}
  \includegraphics[width=2.5cm, height=3.5cm]{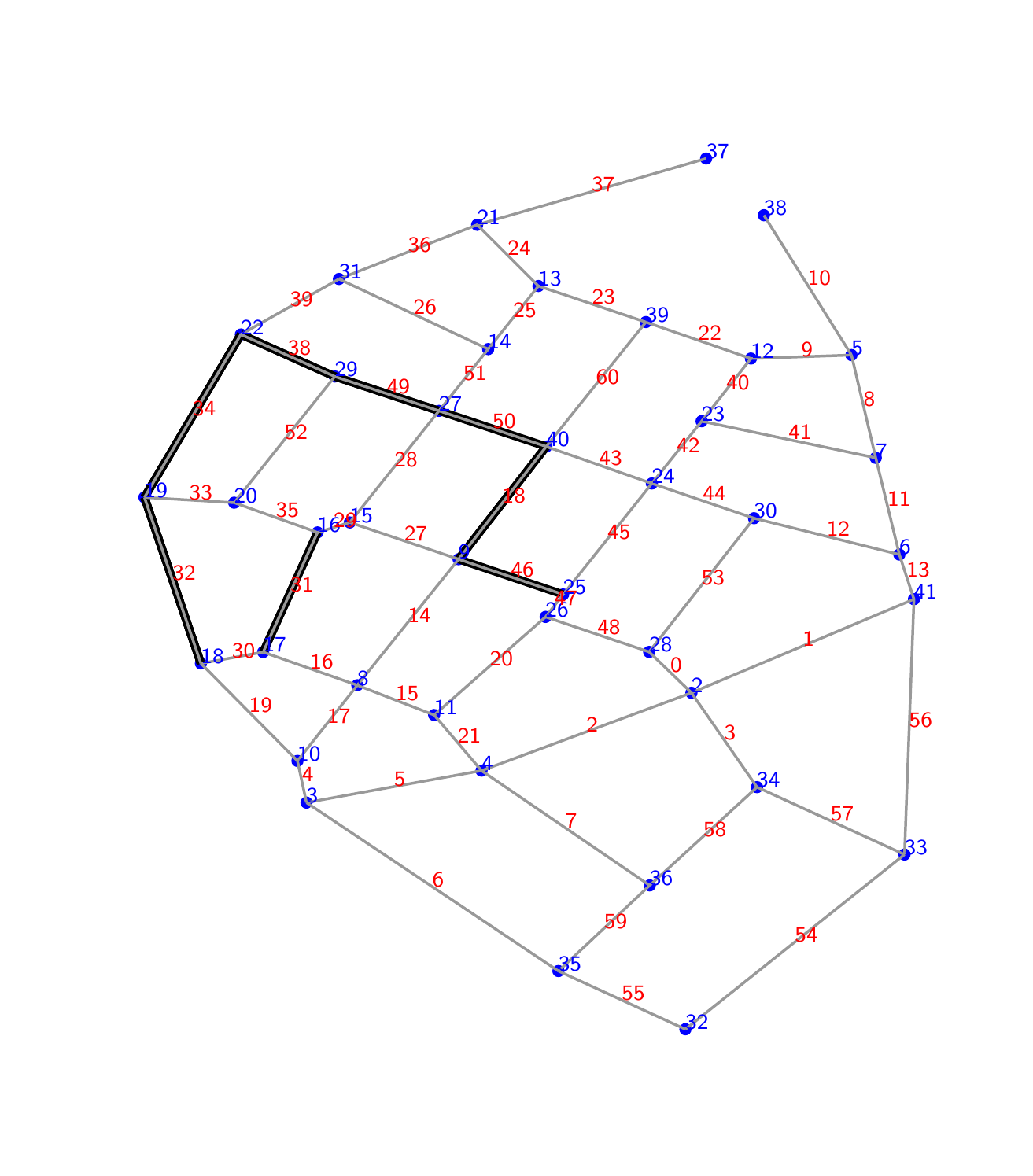}
\end{subfigure}  \hspace{0.5mm}
\begin{subfigure}[t!]{0.15\linewidth}
  \includegraphics[width=2.5cm, height=3.5cm]{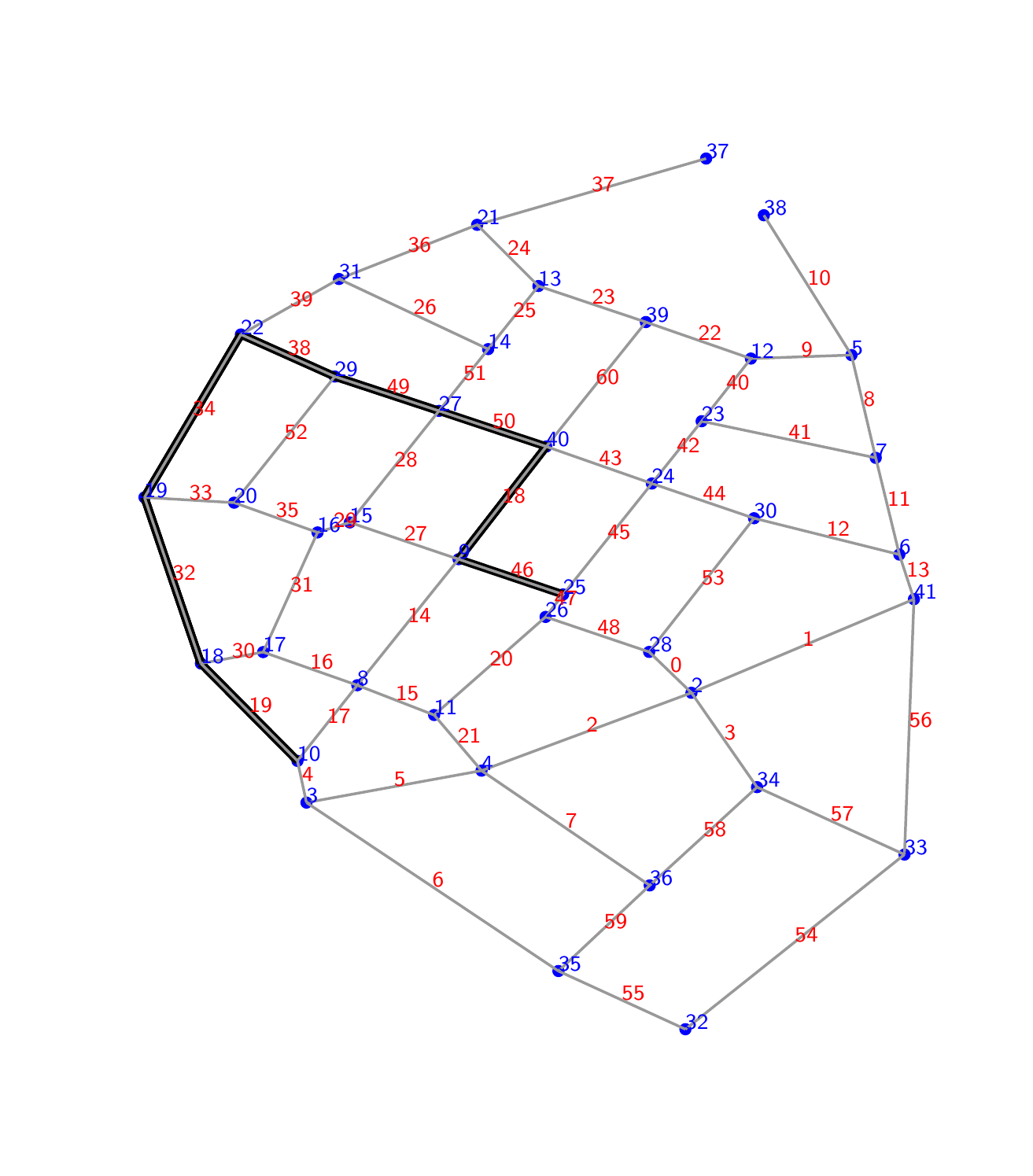}
\end{subfigure}   \hspace{0.5mm}
\begin{subfigure}[t!]{0.15\linewidth}
  \includegraphics[width=2.5cm, height=3.5cm]{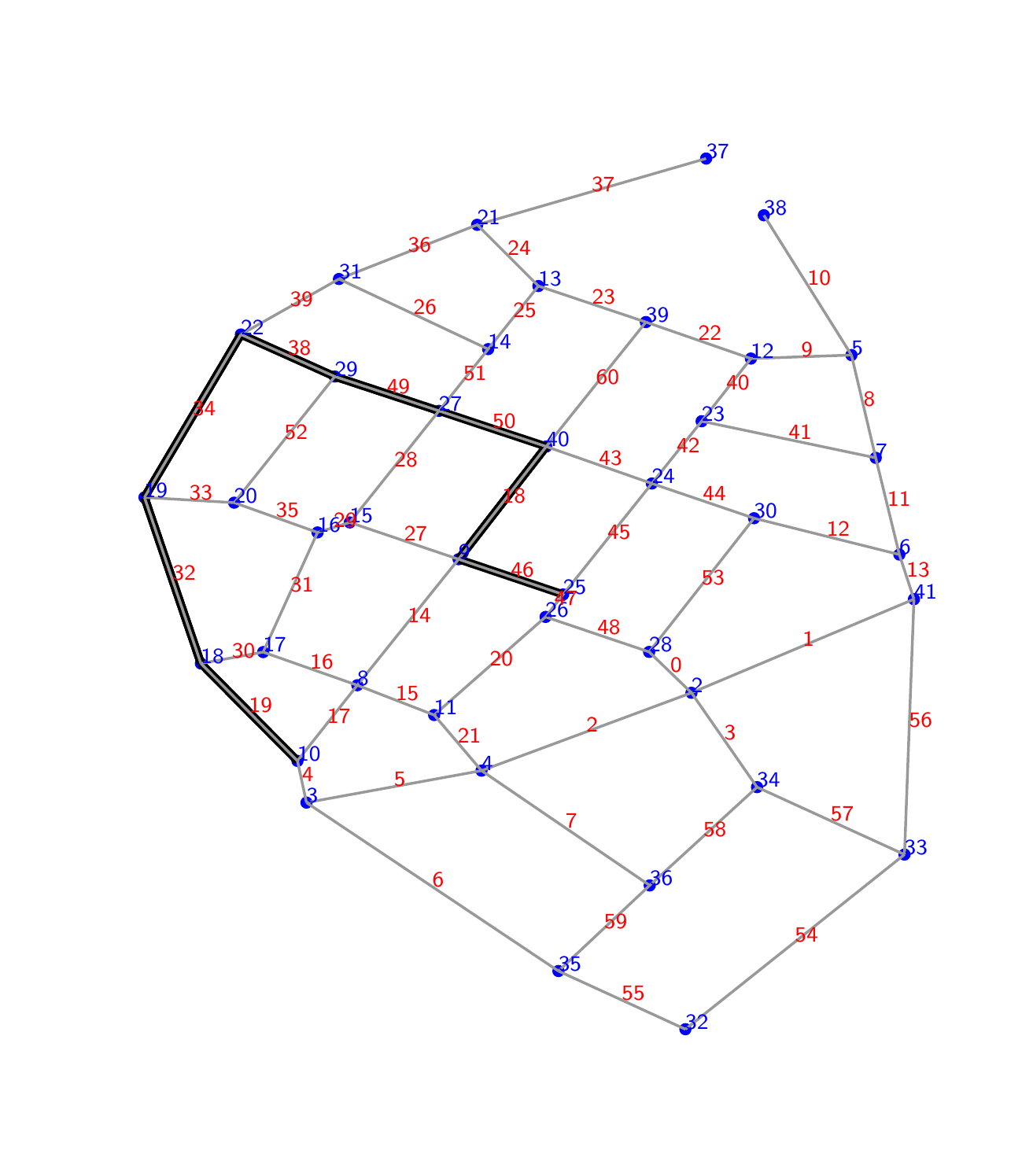}
\end{subfigure}

\vspace{2mm}
\begin{subfigure}[t!]{0.15\linewidth}
  \includegraphics[width=2.5cm, height=3.5cm]{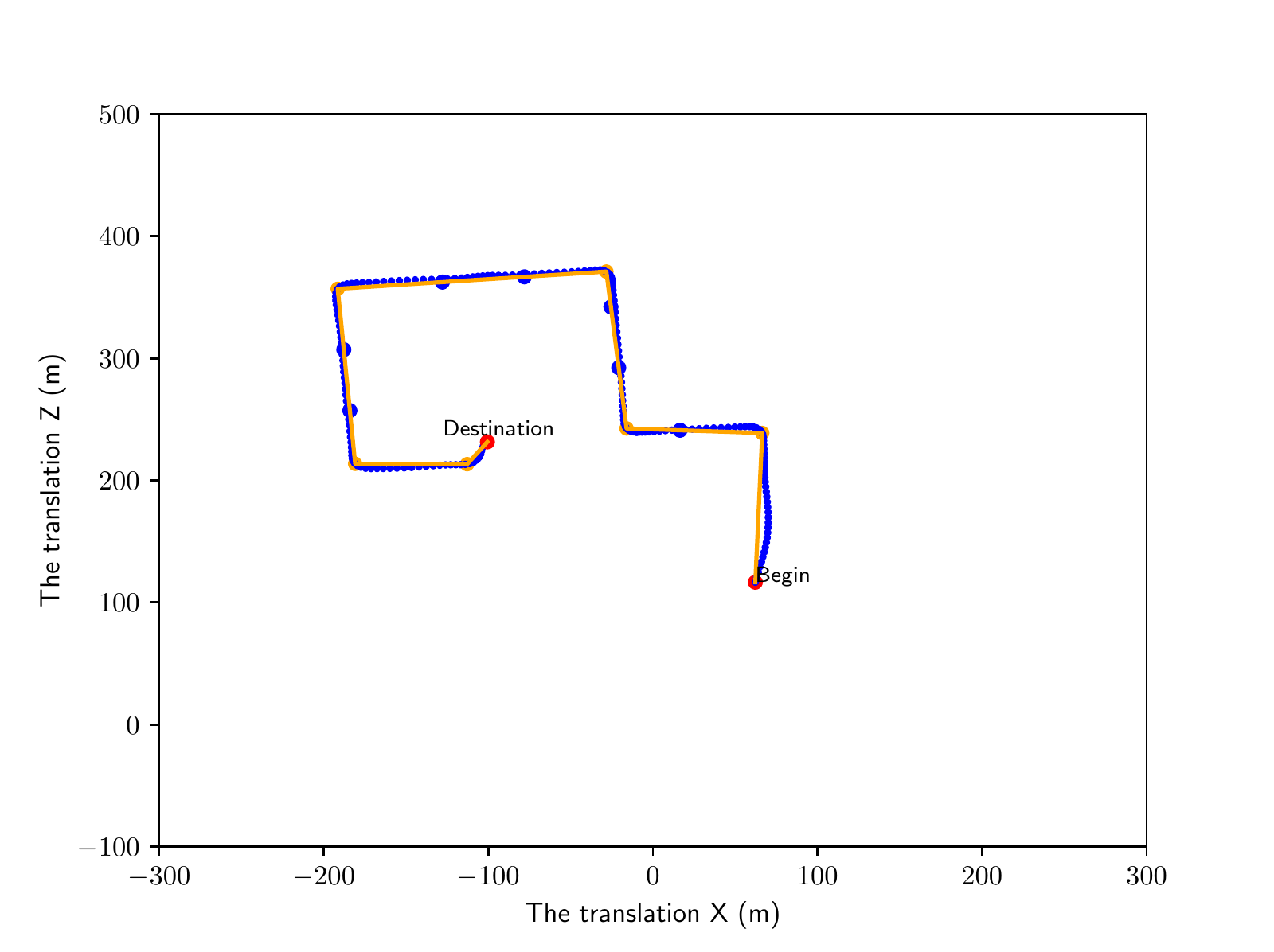}
\end{subfigure}  \hspace{0.5mm}
\begin{subfigure}[t!]{0.15\linewidth}
  \includegraphics[width=2.5cm, height=3.5cm]{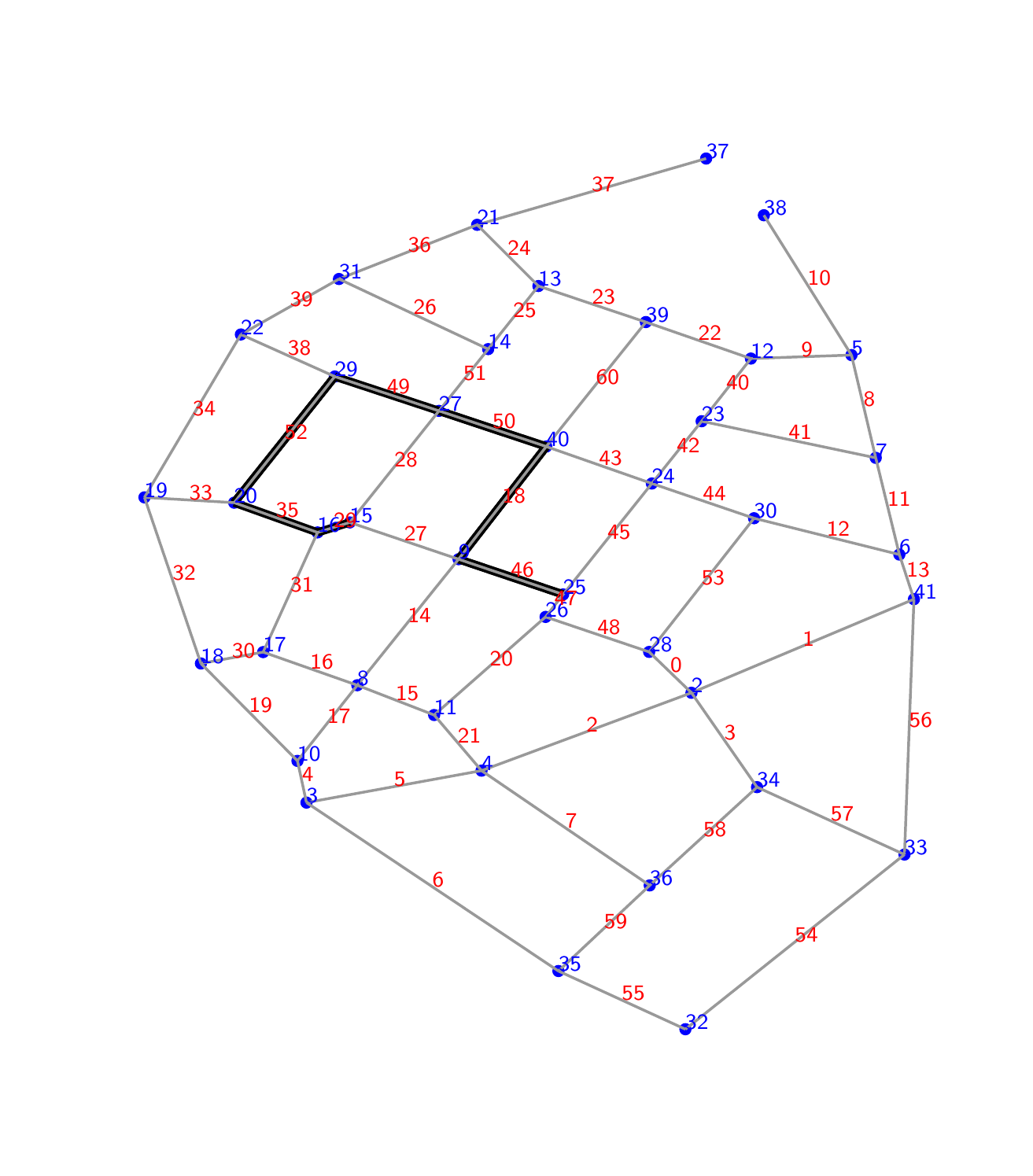}
\end{subfigure}   \hspace{0.5mm}
\begin{subfigure}[t!]{0.15\linewidth}
  \includegraphics[width=2.5cm, height=3.5cm]{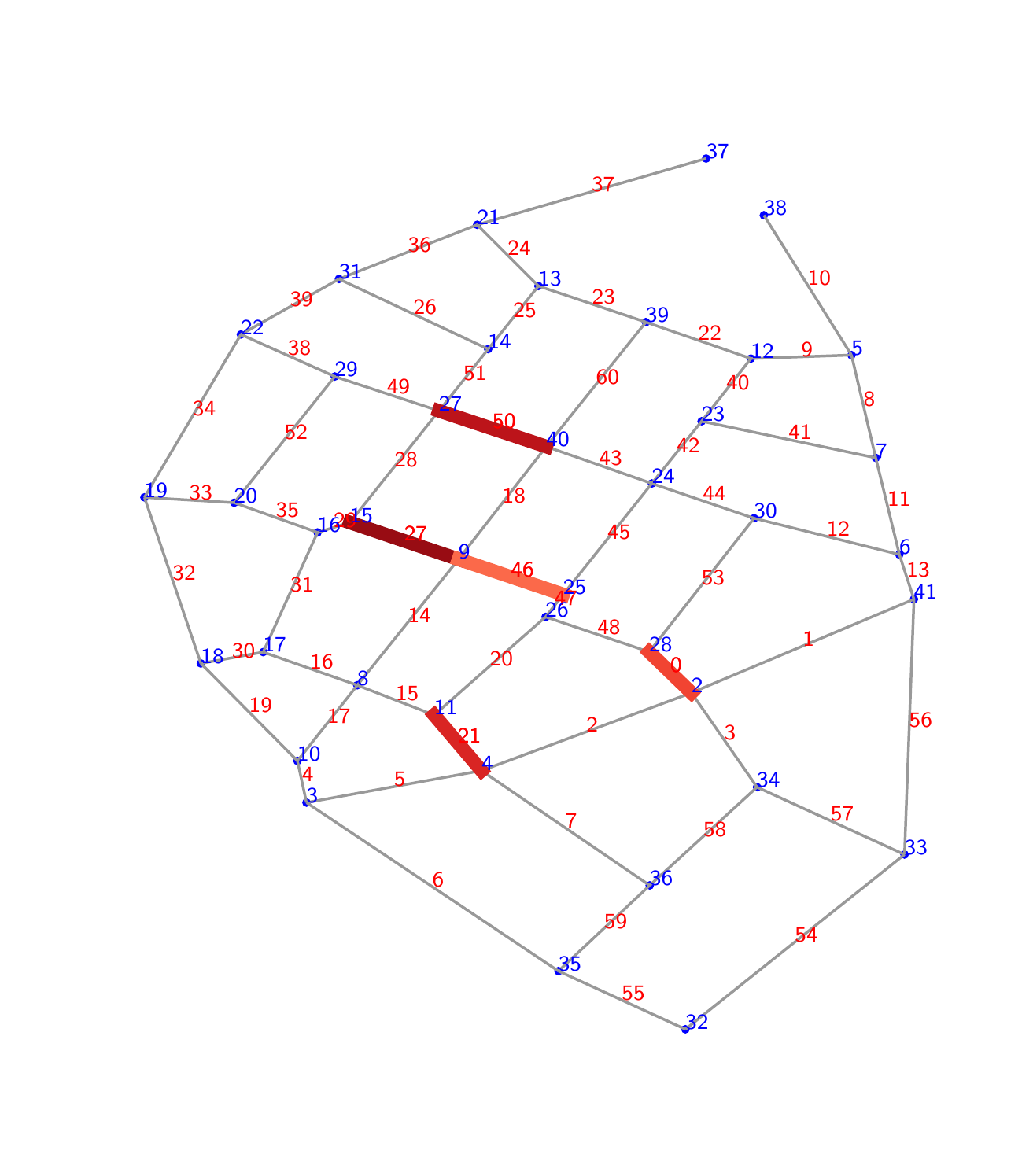}
\end{subfigure}   \hspace{0.5mm}
\begin{subfigure}[t!]{0.15\linewidth}
  \includegraphics[width=2.5cm, height=3.5cm]{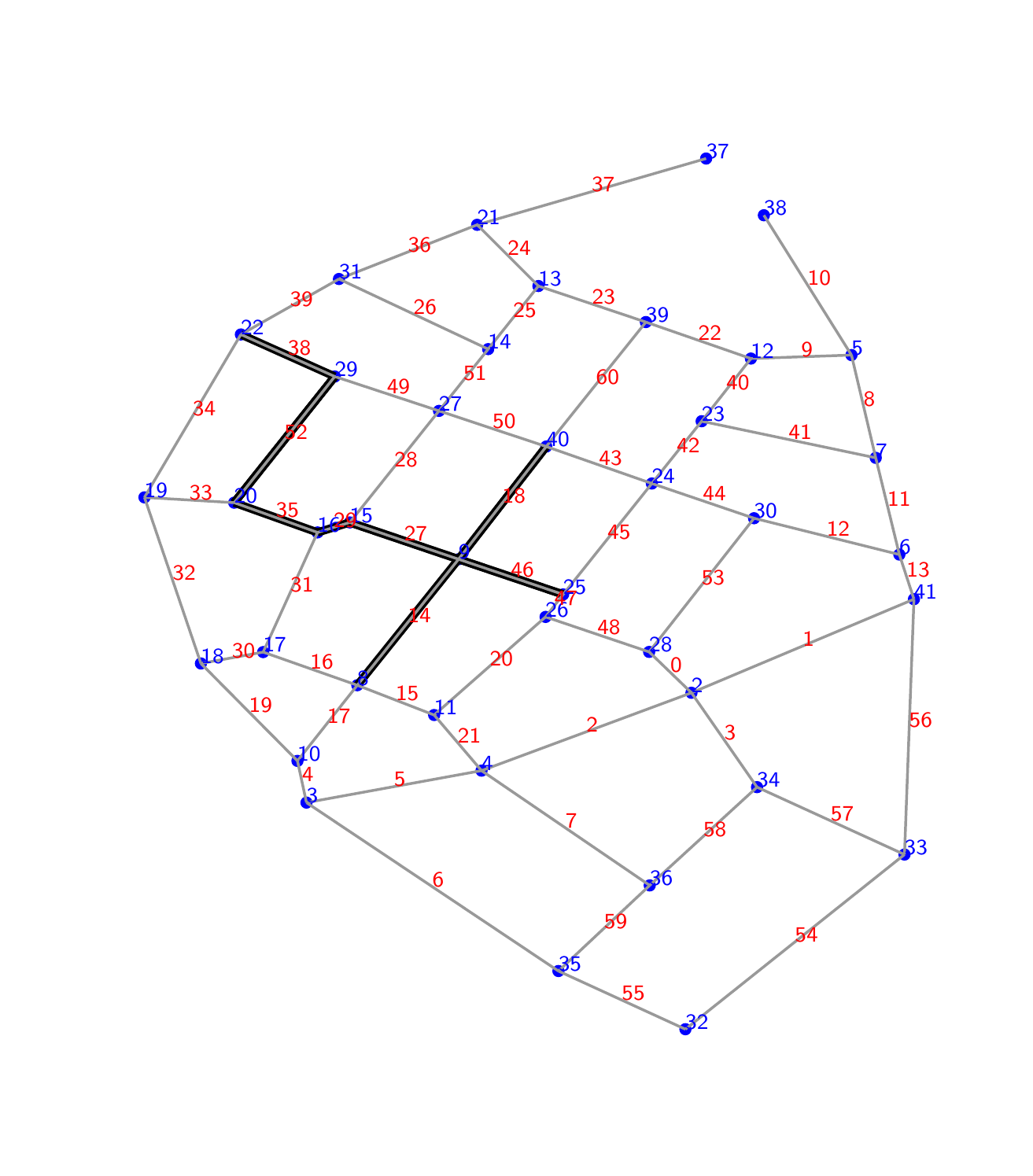}
\end{subfigure}   \hspace{0.5mm}
\begin{subfigure}[t!]{0.15\linewidth}
  \includegraphics[width=2.5cm, height=3.5cm]{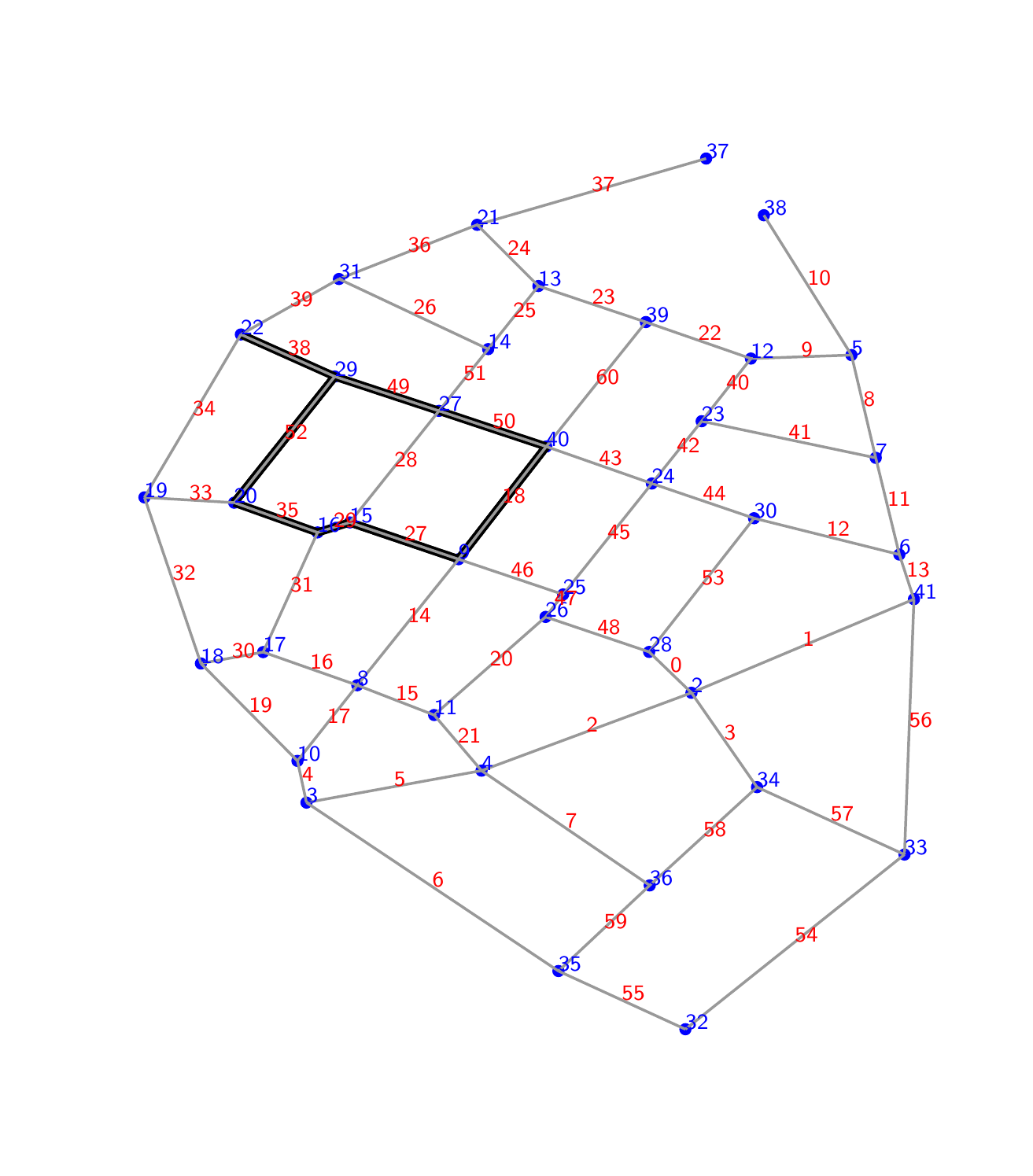}
\end{subfigure}   \hspace{0.5mm}
\begin{subfigure}[t!]{0.15\linewidth}
  \includegraphics[width=2.5cm, height=3.5cm]{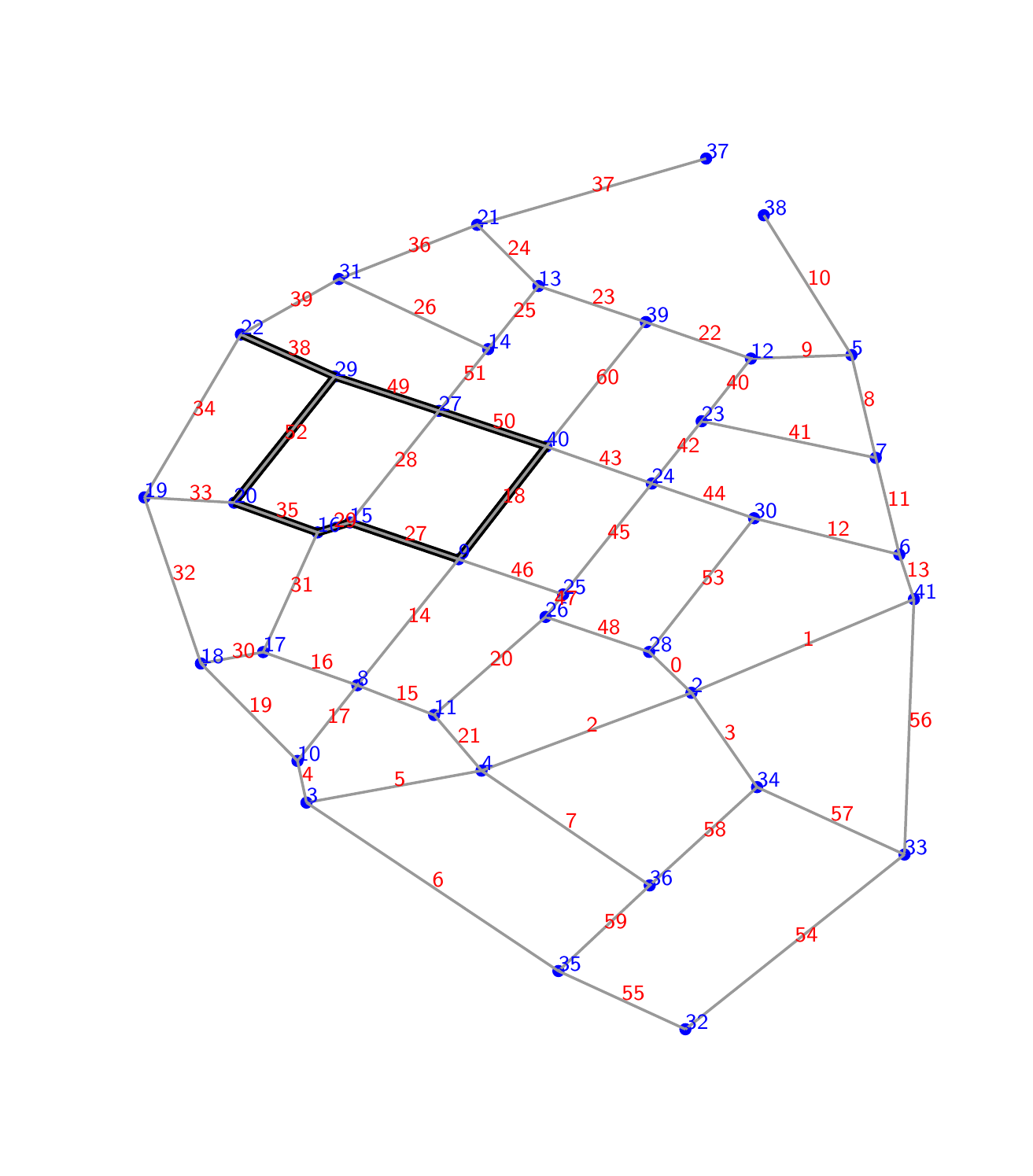}
\end{subfigure}

\vspace{2mm}
\begin{subfigure}[t!]{0.15\linewidth}
  \includegraphics[width=2.5cm, height=3.5cm]{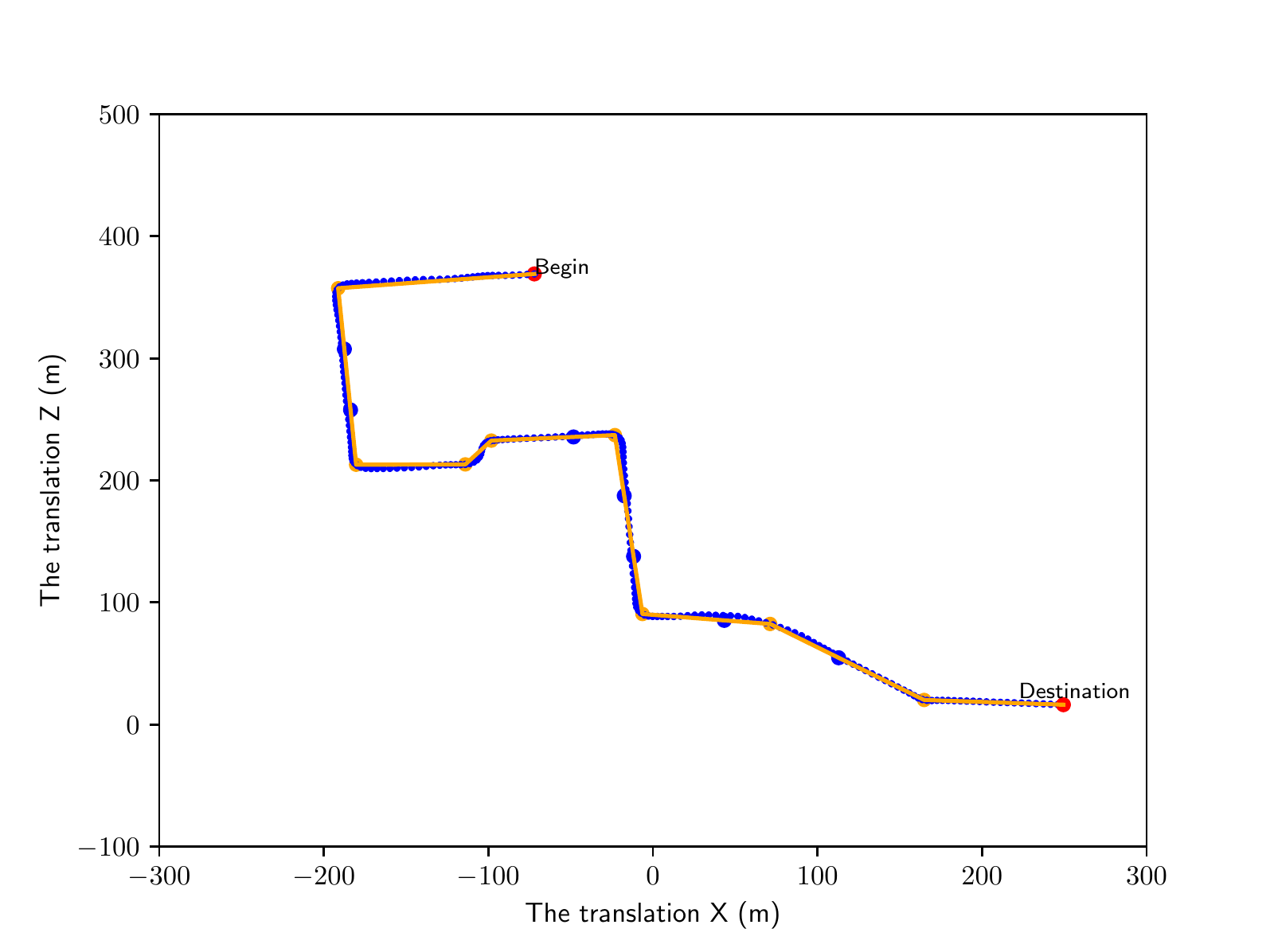}
  \caption{}  
  \label{fig:result_a} 
\end{subfigure}  \hspace{0.5mm}
\begin{subfigure}[t!]{0.15\linewidth}
  \includegraphics[width=2.5cm, height=3.5cm]{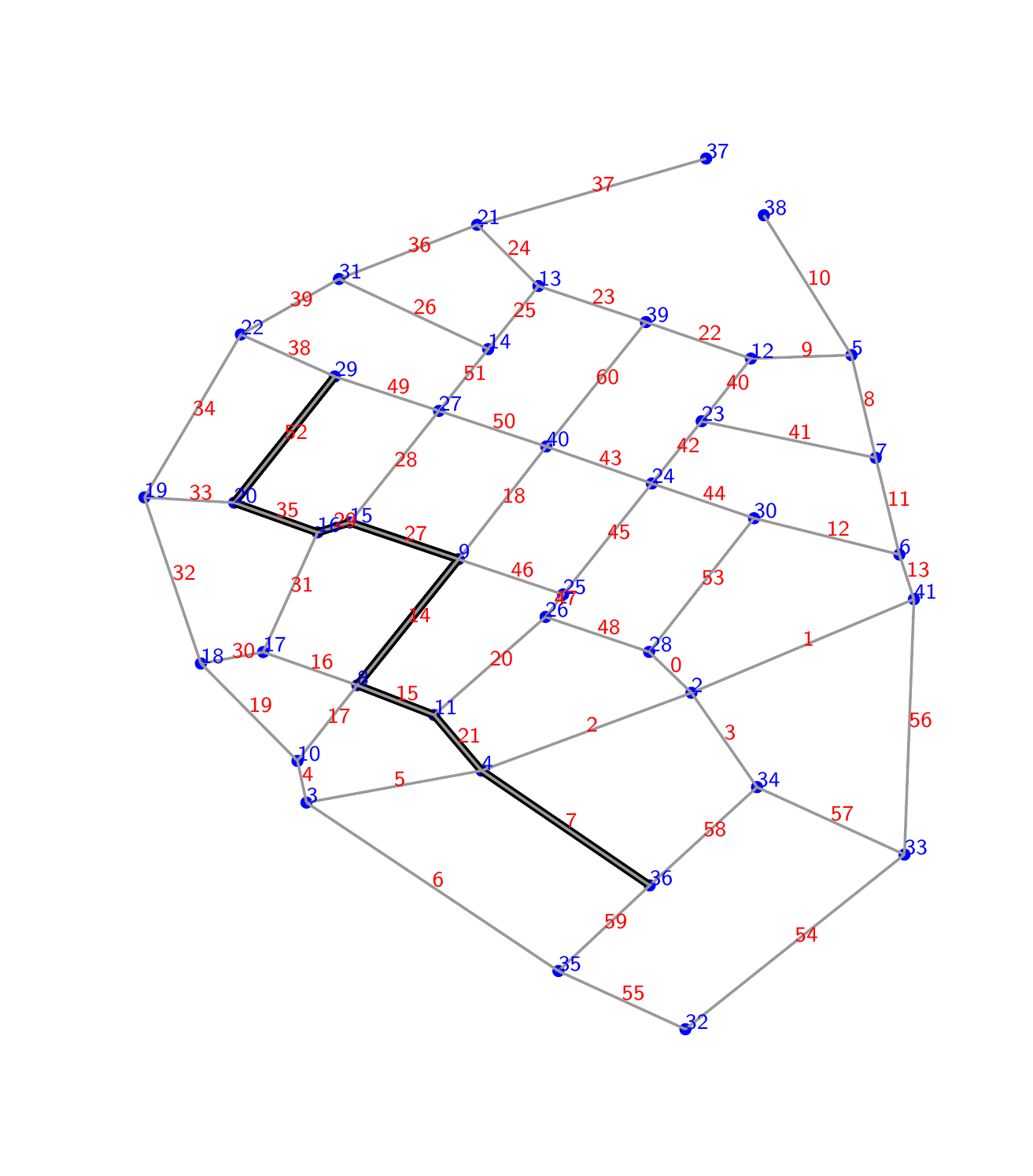}
    \caption{}
  \label{fig:result_b}
\end{subfigure}  \hspace{0.5mm}
\begin{subfigure}[t!]{0.15\linewidth}
  \includegraphics[width=2.5cm, height=3.5cm]{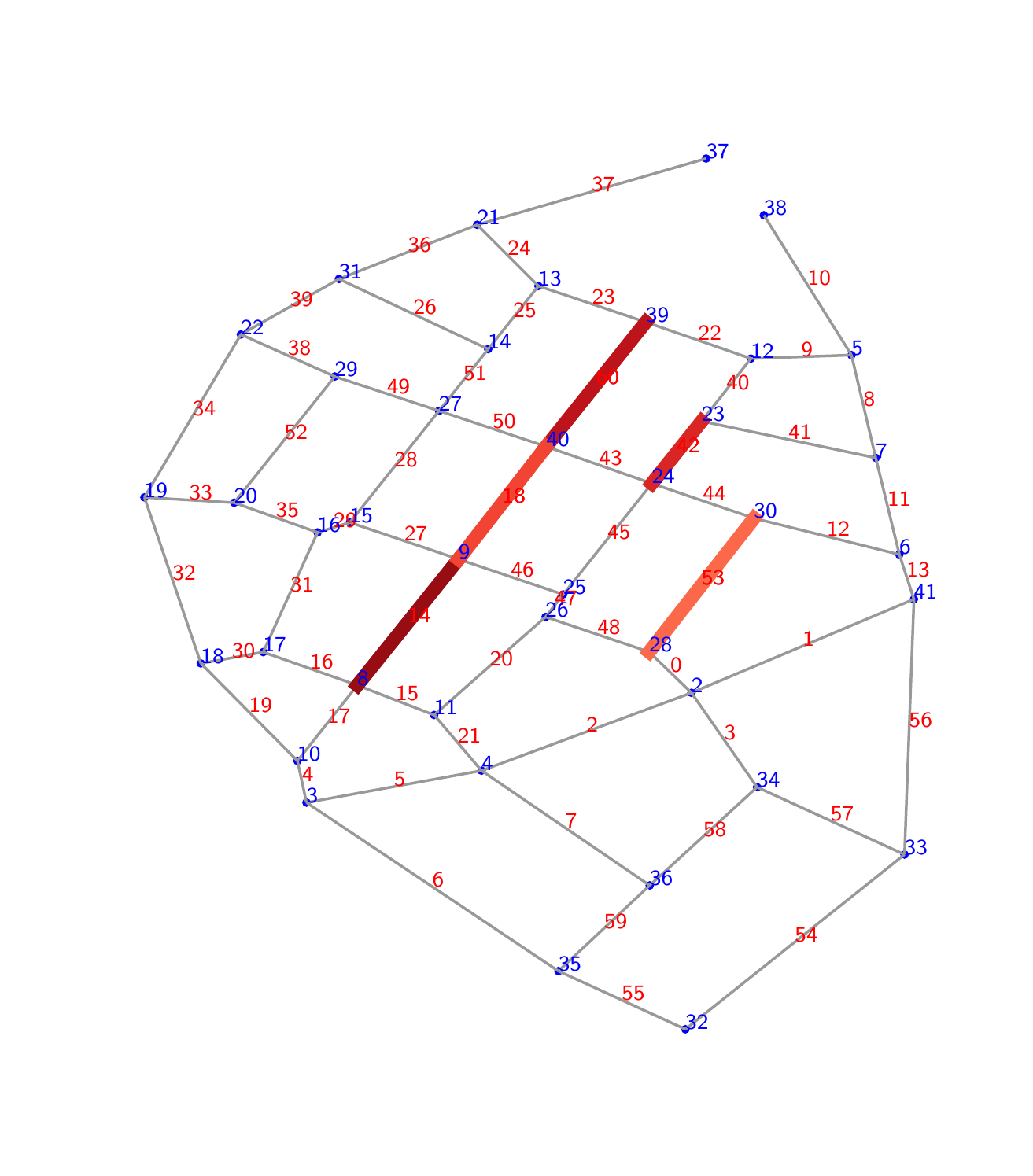}
  \caption{}
  \label{fig:result_c}
\end{subfigure}   \hspace{0.5mm}
\begin{subfigure}[t!]{0.15\linewidth}
  \includegraphics[width=2.5cm, height=3.5cm]{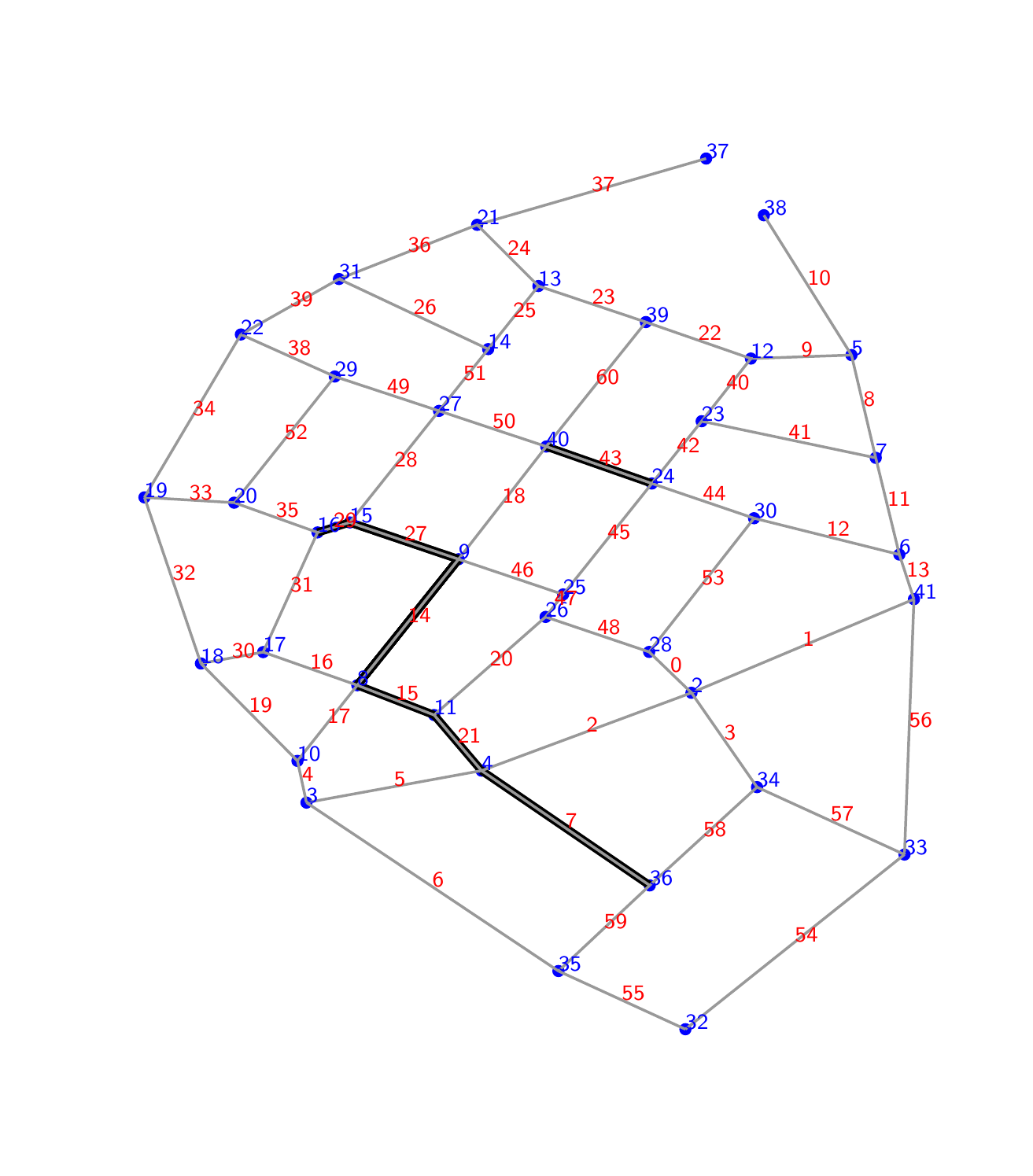}
  \caption{}
  \label{fig:result_d}
\end{subfigure}   \hspace{0.5mm}
\begin{subfigure}[t!]{0.15\linewidth}
  \includegraphics[width=2.5cm, height=3.5cm]{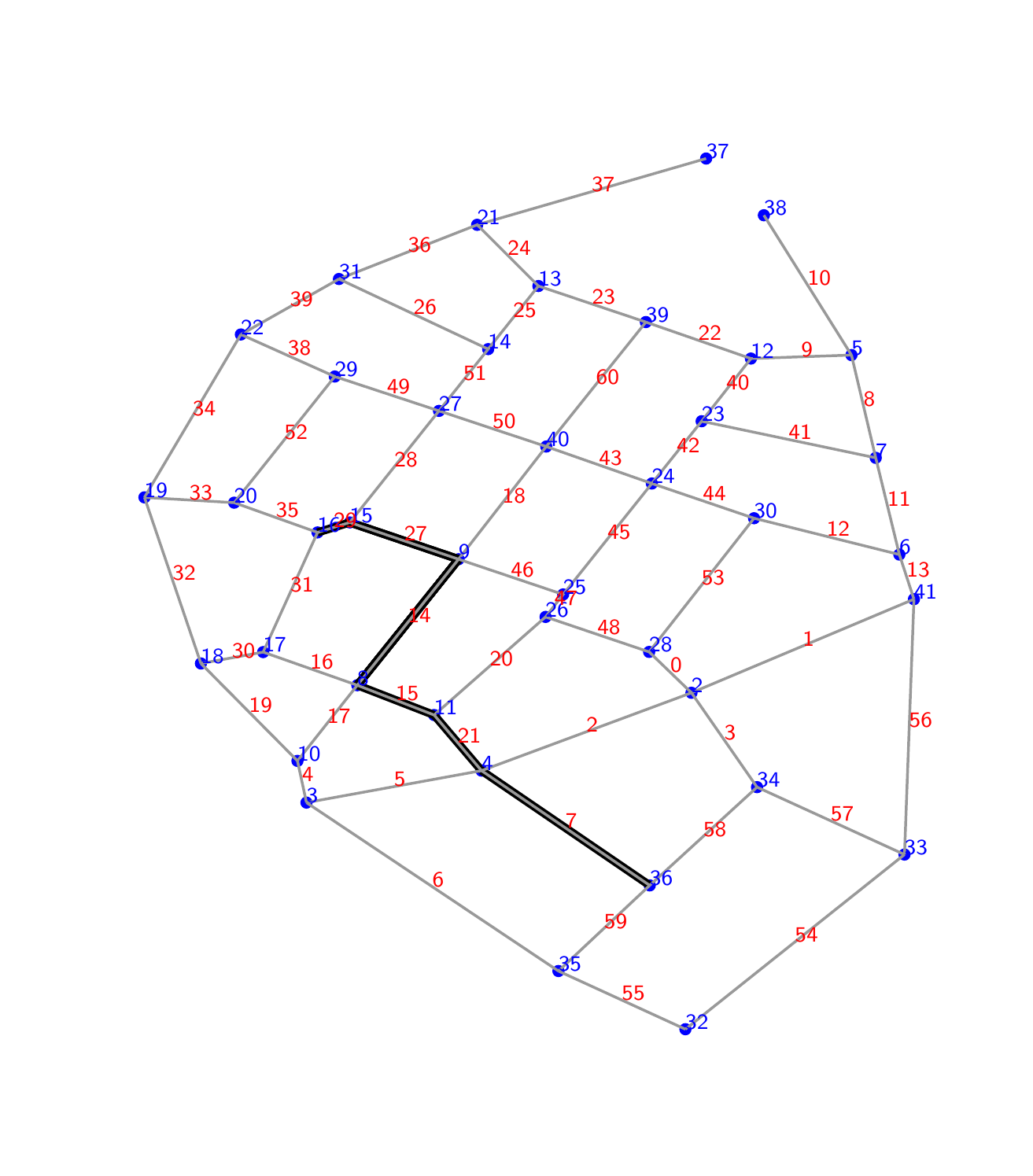}
  \caption{}
  \label{fig:result_e}
\end{subfigure}   \hspace{0.5mm}
\begin{subfigure}[t!]{0.15\linewidth}
  \includegraphics[width=2.5cm, height=3.5cm]{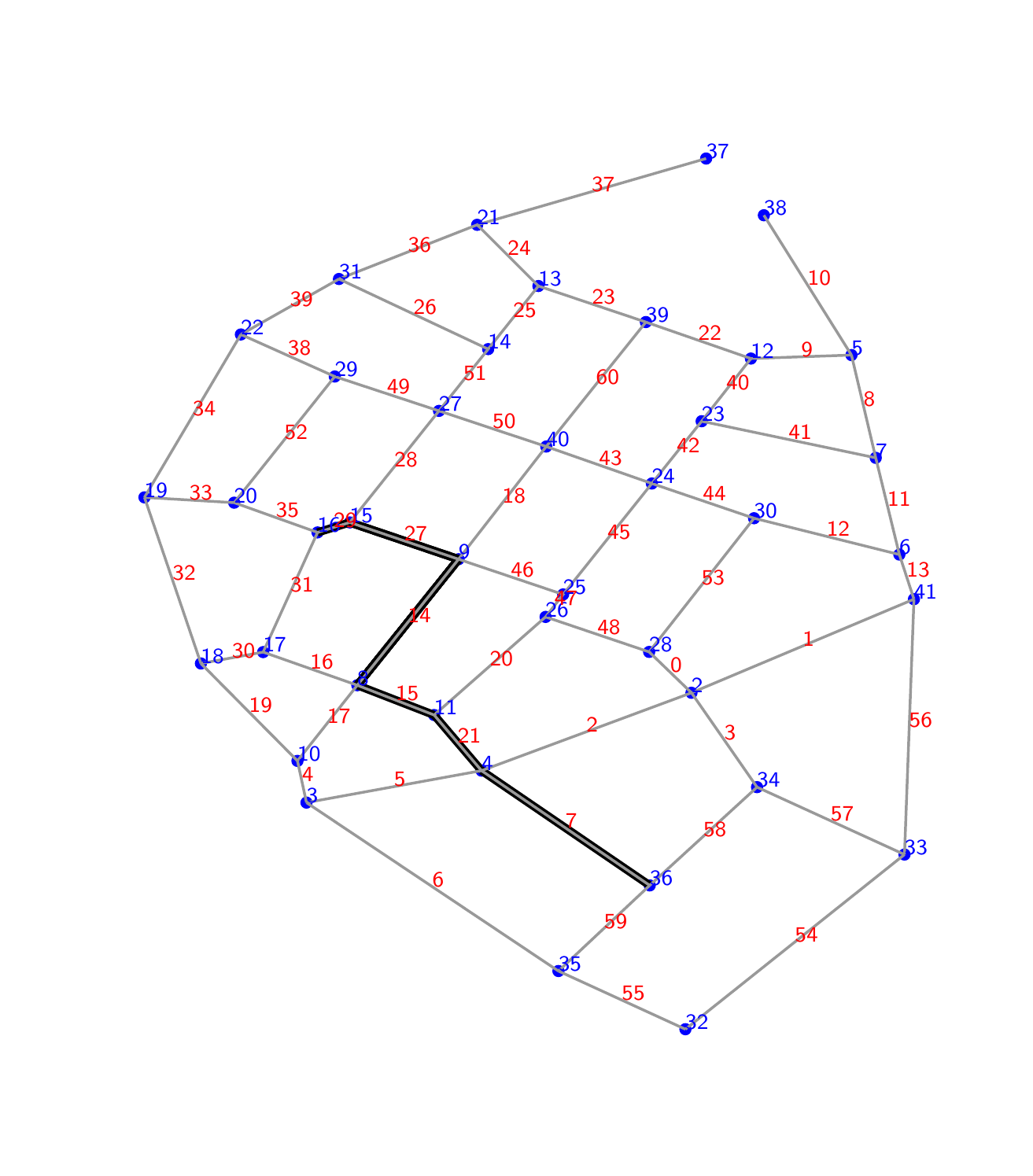}
  \caption{}
  \label{fig:result_f}
\end{subfigure} 

\caption{Each row of results show four different real trajectories from the KITTI dataset. The trajectories are recovered using visual odometry. From first column to last column: (a) vehicle's motion trajectory; blue lines denote visual odometry output and orange lines denote trajectory segments fitted by \cite{douglas1973algorithms}; (b) ground truth edge sequence superimposed on map; (d) predicted edge sequence without temporal consistency; (c) top five edge locations used to start hypotheses for both strategies; (e) temporally consistent localization from strategy 1 and (f) from strategy 2.}
\label{fig:result}
\end{figure*}

To test performance of two hypotheses generation and elimination strategies for temporal consistency, four real trajectories from the KITTI benchmark are selected. Fig. \ref{fig:result} shows these trajectories in column (a) represented by turns with  auxiliary nodes inserted between the turns. The GPS position of the edges superimposed on the map are shown in column (b). This will be used as ground truth comparison. In column (d) of Fig. \ref{fig:result}, we show the predicted edge locations without temporal consistency process which results in disconnected trajectories. The column (c) shows the top five initial edge hypotheses that are used by both strategies to generate hypotheses. Note that, one can also start with all the edges as initial hypotheses and in our tests starting with all edges as initial hypotheses resulted with the same performance as selecting top 5.

As shown in Fig. \ref{fig:result} column (e), we observe that Strategy 1 generates temporally consistent results in case of selecting top 5 or all map edges as initial hypotheses. In contrast, for Strategy 2 (Fig. \ref{fig:result} column (f)), top 5 hypotheses selection in each time also produce satisfying consistent results. 
Both hypotheses generation and elimination strategies obtain similar results for both successful and failure cases 
As shown in Fig. \ref{fig:result}, the first, second and fourth row are able to produce temporally consistent results in the form of edge sequence for both strategies. However, the third row shows the edge sequence is already consistent while it is not the same as ground truth which may be caused by ambiguous trajectories that contain similar distances and turns.  

\section{Conclusion}
\label{sec:con}
In this paper, we propose a novel trajectory learning and topological map-based geolocalization approach, along with a multiple hypotheses generation and elimination strategy for incrementally consistent geolocalization of moving platforms. We formulate trajectory learning process as conditional sequence prediction problem and consistent edge localization prediction as hypotheses generation and elimination problem. We have demonstrated the effectiveness of our approach on a variety of diverse real stereo visual odometry KITTI dataset. In particular, our approach is robust to noise and independent on specific sensor even stereo camera used in this paper. In the future, we will further explore the more efficient temporally consistent localization to improve the final edge sequence and intend to also integrate the temporal consistency into neural network instead only a post-processing step. Geolocalization in a larger map area is also challenging in the future work.

\bibliographystyle{ieeetr}
\bibliography{references}

\end{document}